%% file: main.tex
\documentclass{article}

    \PassOptionsToPackage{numbers, compress}{natbib}

     \usepackage[preprint]{neurips_2020}

\usepackage[utf8]{inputenc} 
\usepackage[T1]{fontenc}    
\usepackage{hyperref}       
\usepackage{url}            
\usepackage{booktabs}       
\usepackage{amsfonts}       
\usepackage{nicefrac}       
\usepackage{microtype}      
\usepackage{chngcntr}
\usepackage{times}  
\usepackage{helvet} 
\usepackage{courier}  
\usepackage{multicol}
\usepackage{authblk}
\usepackage{breakcites}
\usepackage{placeins}
\usepackage{subcaption}
\usepackage{graphicx} 
\usepackage{graphicx}  
\input{math_commands.tex}

\usepackage[utf8]{inputenc} 
\usepackage[T1]{fontenc}    
\usepackage{hyperref}       
\usepackage{epsfig}
\usepackage{graphicx}
\usepackage{booktabs}       
\usepackage{amsfonts}       
\usepackage{amsmath}
\usepackage{amssymb}
\usepackage{nicefrac}       
\usepackage{microtype}      
\usepackage{multirow}
\usepackage{xspace}
\usepackage{array}
\usepackage{afterpage}
\usepackage{textcomp}

\usepackage{siunitx}
\usepackage{booktabs}
\usepackage{arydshln}   
\usepackage{color}
\newcommand{\mli}[1]{\mathit{#1}}

\newcommand{\supl}{Supp. Mat. }

\renewcommand{\i}{$(\textbf{i})$ }
\newcommand{\ii}{$(\textbf{ii})$ }
\newcommand{\iii}{$(\textbf{iii})$ }
\newcommand{\iv}{$(\textbf{iv})$ }
\renewcommand{\v}{$(\textbf{v})$ }

\usepackage{graphicx}
\title{CogMol: Target-Specific and Selective Drug Design for COVID-19 Using Deep Generative Models}

\author{%
Vijil Chenthamarakshan, Payel Das, Samuel C. Hoffman, Hendrik Strobelt$^{\dagger}$, Inkit Padhi, Kar Wai Lim$^{\ast}$, Benjamin Hoover$^{\dagger}$, Matteo Manica$^{\ddagger}$, Jannis Born$^{\ddagger}$, Teodoro Laino$^{\ddagger}$, Aleksandra Mojsilovic\\
  IBM Research, Yorktown Heights, New York; $^{\ast}$IBM Research, Singapore\\
  $^{\dagger}$IBM Research, MIT-IBM Watson AI Lab, Cambridge; $^{\ddagger}$IBM Research Europe \\
  \texttt{\{ecvijil,daspa,aleksand\}@us.ibm.com, \{shoffman,hendrik.strobelt\}@ibm.com, \{inkpad,kar.wai.lim,benjamin.hoover\}@ibm.com, \{tte,jab,teo\}@zurich.ibm.com} \\
}

\begin{document}

\maketitle

\begin{abstract}
The novel nature of SARS-CoV-2 calls for the development of efficient de novo drug design approaches. In this study, we propose an end-to-end framework, named CogMol (Controlled Generation of Molecules), for designing new drug-like small molecules targeting novel viral proteins with high affinity and off-target selectivity. CogMol combines adaptive pre-training of a molecular SMILES Variational Autoencoder (VAE) and an efficient multi-attribute controlled sampling scheme that uses guidance from attribute predictors trained on latent features. To generate novel and optimal drug-like molecules for unseen viral targets, CogMol leverages a protein-molecule binding affinity predictor that is trained using SMILES VAE embeddings and protein sequence embeddings learned unsupervised from a large corpus.
We applied the CogMol framework to three SARS-CoV-2 target proteins: main protease, receptor-binding domain of the spike protein, and non-structural protein 9 replicase. The generated candidates are novel at both the molecular and chemical scaffold levels when compared to the training data. CogMol also includes \insilico screening for assessing  toxicity of parent molecules and their metabolites with a multi-task toxicity classifier, synthetic feasibility with a chemical retrosynthesis predictor, and target structure binding with docking simulations.
Docking reveals favorable binding of generated molecules to the target protein structure, where 87--95\% of high affinity molecules showed docking free energy $<$ -6 kcal/mol. When compared to approved drugs, the majority of designed compounds show low parent molecule and metabolite toxicity and high synthetic feasibility. In summary, CogMol can handle multi-constraint design of synthesizable, low-toxic, drug-like molecules with high target specificity and selectivity, even to novel protein target sequences, and does not need target-dependent fine-tuning of the framework or target structure information.
\end{abstract}

\section{Introduction}


Generating novel drug molecules is a daunting task that aims to create new molecules (or optimize known molecules) with multiple desirable properties that often compete and tightly interact with each other. For example, optimal drug molecules should have binding affinity to the target protein of interest (target specificity), low binding affinity to other targets (off-target selectivity), be easy to synthesize, and also exhibit high drug likeliness (QED). This makes drug discovery a costly (2-3 billion USD) and time-consuming process (more than a decade) with a low success rate ($<$ 10\%) \cite{matthews2016omics}.


Traditional \textit{in silico} molecule design and screening rely on rational design methods that need physics-based simulations, heuristic search algorithms, and considerable domain knowledge.
However, optimizing over the discrete, unstructured and sparse molecular space remains an intrinsically difficult challenge. 
Therefore, there is much interest in developing automated machine learning techniques to efficiently discover sizeable numbers of plausible, diverse and novel candidate molecules in the vast ($10^{23}\textit{-}10^{60}$) space of molecules \cite{polishchuk2013estimation}. Bayesian optimization, Reinforcement Learning, and gradient-based optimization methods have been proposed for automating drug molecule design with desired properties (e.g., high drug-likeliness, synthetic accessibility, or solubility) \cite{gomez2018automatic, Zhavoronkov2019natbio, zhou2019optimization}. 
These methods either optimize directly on the high-dimensional input space or on the low dimensional representation learned using a latent variable model such as a probabilistic autoencoder. 

%

One crucial aspect of designing drug candidates is to account for the right context, e.g., protein, gene, metabolic or disease pathway  information.
For example, in target protein-specific drug design, the goal is to generate molecules with high binding affinity to a specific target protein. This requires fine-tuning a generative model on a small library of ligands to enable target-specific sampling.
For novel or unrelated proteins, such as the SARS-CoV-2 viral proteins involved in the recent COVID-19 pandemic, binding affinity data is unavailable. 
At the same time, these novel target proteins are not related to the proteins in  existing binding affinity databases.
Thus, handling novel targets in the current generative frameworks becomes non-trivial. 

Designing drug candidates for novel targets gets even more challenging, as the drug designed for the novel target can bind to other undesired targets. 
Small molecule drugs have been shown to bind on average to a minimum of 6-11 distinct targets in addition to their intended target \cite{peon2017predicting}. 
This molecular ``promiscuity'' of drugs causes unintended therapeutic effects or multiple drug–target interactions leading to off-target toxicities and decreased effectiveness \cite{cheng2019network, miljkovic2018data}.
Accounting for this important aspect of off-target selectivity 
becomes non-trivial if the generative model is trained only on a small ligand library optimized for a single target or only on good binder molecules for a limited set of targets. 

\section{CogMol - Molecule Generation Pipeline}

\begin{figure*}[tb]
    \centering
    \includegraphics[width=0.95\textwidth]{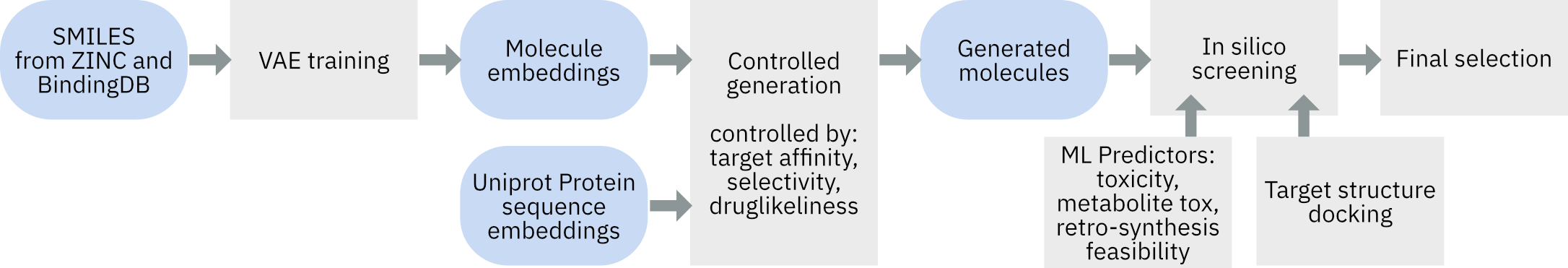}
    \vspace{1pt}
    \caption{Workflow of the drug candidate generation pipeline}
    \label{fig:workflow}
    \vspace{-8pt}
\end{figure*}

To address the challenges stated above,  
we propose an alternative method named \textbf{Co}ntrolled \textbf{G}eneration of \textbf{Mol}ecules (CogMol) for designing small molecule drugs in a context, \textit{i.e.} target protein in this study, -dependent manner. CogMol accounts for both target specificity and selectivity, even for novel or low-coverage target sequences. 
As depicted in Figure~\ref{fig:workflow}, CogMol includes the following components:

    \textbf{1.} A Variational Autoencoder (VAE), first trained unsupervised and then jointly with a set of attribute regressors (QED and Synthetic Accessibility, SA), that learns a disentangled latent space of the molecules using the SMILES representation.\\ 
    \textbf{2.} A protein-molecule binding affinity regressor trained on the VAE latent features of molecules and protein sequence embeddings trained on a large unlabeled corpus, which is used for estimating target specificity and off-target selectivity.\\ 
    \textbf{3.} An efficient sampling scheme to generate molecules with desired attributes from the model of the VAE latent space, using guidance from a set of attribute (affinity, selectivity, QED) predictors. 

Instead of training the binding affinity regressor on the  sequence embeddings of a few thousand target proteins reported in the binding affinity database, CogMol uses pre-trained protein sequence embeddings, \cite{Alley_2019} learned on an unlabeled corpus of 24M Uniprot protein sequences from UniRef50 database, to train the affinity predictor.
Since these pre-trained protein embeddings are capable of better capturing sequence, structural, and functional relationships \cite{Alley_2019, rao2019evaluating}, using them in CogMol allows targeted generation of molecules even for new/unseen targets and does not require model retraining for every individual target. 
Finally, CogMol proposes an efficient way of modeling off-target selectivity and using this as a control for targeted generation, leveraging the same trained protein-ligand binding affinity predictor.  

CogMol is also empowered with an \insilico screening protocol for generated molecules, which involves: \i toxicity prediction on several \emph{in vitro} and clinical endpoints for parent molecules and their predicted metabolites using a multi-task deep learning-based classifier, 
\ii synthetic feasibility prediction using a chemical retrosynthesis predictor; and \iii blind docking simulations to estimate binding of the generated molecules to the target protein structure. We hope that accounting for multiple important  properties that play a role in the efficacy or viability of a drug such as target affinity, off-target selectivity, toxicity of parent molecules and their metabolites, and synthetic practicality,   within an AI framework will help the \insilico drug design process to be faster and less costly, leading to shorter discovery pipelines with high success rate. 

\textbf{CogMol for COVID-19 Antiviral Molecule Design.}
Given the urgency with the ongoing COVID-19 pandemic, we apply CogMol to generate candidate molecules that bind to three relevant target proteins of the novel SARS-CoV-2 virus, namely NSP9 Replicase (NSP9), Main Protease (M\textsuperscript{pro}), and the Receptor-Binding Domain (RBD) of the SARS-CoV-2 S protein, with high affinity.
Note, these targets are not present in the binding affinity database, and both NSP9 and RBD are more novel than M\textsuperscript{pro} (See \supl\ref{SI:targets}). We also used CogMol to generate molecules for human Histone deacetylase 1 (HDAC1) protein implicated in cancer, for which the number of molecules with desired criteria in the training database is very low. 

Our contributions in this work are: \i An end-to-end framework for drug-like small molecule design that accounts for multiple relevant and critical factors such as target affinity, off-target selectivity, toxicity of the parent molecules and their metabolites across different endpoints, target structure binding, and synthetic practicality. \ii This is, to our knowledge, the first deep generative  approach that generates novel, specific, and selective drug-like small molecules for a \textit{unseen} target sequence without model retraining.
\iii A system capable of generating drug-like molecules with high target affinity and selectivity for selected targets that are either relatively novel or have a low ligand coverage. 
\iv Although our framework learns from 1D protein sequence information only, generated molecules are still capable of binding to the druggable binding pockets of the 3D target structure with favorable binding free energy (BFE). 
\v For three novel and very relevant COVID-19 targets (as well as for a cancer target with low coverage of optimal ligands), we are able to identify a set of generated novel and optimal drug-like molecules with high target affinity and selectivity, that binds well to the target structure, is synthetically practical and has low predicted parent and metabolite toxicity with respect to FDA-approved drugs.

\section{Related Work}
Earlier approaches to generate molecules involved recurrent neural networks (RNN) \cite{segler2017generating, gupta2018generative}, whereas recent works employ deep generative frameworks, such as the Variational Autoencoder (VAE) \cite{gomez2018automatic, blaschke2018application, kang2018conditional, lim2018molecular} and the Generative Adversarial Network (GAN) \cite{guimaraes2017objective, de2018molgan}. 
Most of those works employ SMILES representation. Generating syntactically valid molecules under SMILES grammar is challenging and there have been attempts to ensure validity via constraints~\cite{kusner2017grammar, dai2018syntax}.
Recently, there has been increasing interest in molecular graph-based generative methods \cite{li2018learning, li2018multi, simonovsky2018graphvae, samanta2019nevae,ma2018constrained, kajino2018molecular}. 
Unfortunately, 
graph-based models are not permutation-invariant of their node labels, the training has a quadratic complexity concerning the number of nodes, and generating semantically valid graphs is challenging.
\cite{jin2018junction} is considered as a state-of-the-art architecture in this context,  which represents a molecular graph as fragments 
connected in a tree structure. 

For targeted generation of molecules with a set of desired attributes, Reinforcement Learning (RL) and Bayesian Optimization (BO) methods have often been employed on top of a SMILES- or Graph-based molecule generator  \cite{popova2018deep, olivecrona2017molecular,jaques2017sequence, putin2018reinforced, Zhavoronkov2019natbio, gomez2018automatic, born2020paccmann, zhou2019optimization}, but typically incur high computational cost. 
Semi-supervised learning has also been used  \cite{lim2018molecular, kang2018conditional, li2018multi}, which involves optimizing complicated loss objectives. 
CogMol is instead inspired by the Controlled Latent attribute Space Sampling (CLaSS) method \cite{das2020science}. 
Our proposed methodology aims at computationally efficient targeted generation with multiple constraints from the latent space, requiring minimal model training and no supervision.

To generate drugs specific to a particular target, generative models in existing works \cite{Zhavoronkov2019natbio, li2018multi, skalic2019target} are typically fine-tuned on the subset of molecules that bind to that specific target sequence or structure and hence cannot generalize to unseen targets.
Recently, target-specific de novo drug design has been defined as a translation problem between amino acid ``language'' and molecular SMILES \cite{grechishnikova2019transformer}, where the latent code $z$ of a protein is considered as a ``context'' to generate a binding ligand.
However, the protein embeddings are learned only from the $\sim$1100 human protein sequences captured in BindingDB, which limits the model's generalization capabilities.
In contrast, CogMol uses  protein embeddings from Unirep trained on an  unsupervised corpus of $\sim$24 million UniRef50 amino acid sequences~\cite{Alley_2019}. This approach has been demonstrated to improve performance in downstream prediction tasks~\cite{rao2019evaluating, Alley_2019} as well as in generative modeling~\cite{das2020science}.

\noindent

\section{Model and Methods}
\textbf{Data.}
We used the Moses benchmarking dataset \cite{polykovskiy2018molecular} for the unsupervised VAE training, which include 1.6M molecules in the training set and 176K molecules in the test and scaffold test sets respectively from the ZINC database \cite{irwin2005zinc}. 

For target-specific compound design, we used a curated IC50-labeled compound-protein binding data from BindingDB \cite{gilson2015bindingdb}, as reported in DeepAffinity \cite{karimi2018deepaffinity}. The DeepAffinity models use a separate held-out set with four different protein classes to test the generalizability  of their predictor. 
Since our objective is to build the best binding affinity (pIC50) regression model using available data, we also added the four excluded classes into our training data.

\textbf{Variational Autoencoder for Molecule Generation.}
A Variational Autoencoder (VAE) ~\cite{kingma2013auto} frames an autoencoder in a probabilistic formalism that constrains the expressivity of the latent space, $z$. Each sample defines an encoding distribution $q_\phi(z|x)$ and for each sample, this encoder distribution is constrained to be close to a simple prior distribution $p(z)$.
We consider the case of the encoder specifying a diagonal Gaussian distribution only, i.e. $q_\phi(z|x) = N(z; \mu(x), \Sigma(x))$ with $\Sigma(x) = \text{diag}[\exp(\log(\sigma_d^2)(x)))]$.
The encoder neural network produces the log variances $\log(\sigma_d^2)(x)$. The marginal posterior is $q_\phi(\rvz) = \frac{1}{N} \sum_{j=1}^N q_\phi(\rvz|\rvx_j)$.
The standard VAE objective is defined as follows (where $\KL$ is the Kullback-Leibler divergence),
$
     \mathcal{L}_{\text{VAE}}(\theta, \phi) = \E_{p(x)} \left\{
    \E_{q_\phi(z|x)}[\log p_\theta(x|z)] 
    - \KL(q_\phi(z|x) || p(z))
    \right\}
$.

We also jointly trained two property predictors, one for QED  and one for SA, each parameterized by a feed-forward network, along with the VAE, to predict $y(x)$ from the latent embedding of $x$. As shown in \supl\ref{SI:hdca1}, the BindingDB molecules have a different distribution of QED when compared to molecules in ZINC. To better reflect this diversity in the latent embeddings of the VAE,  
we continued training of the VAE model with QED and SA predictors on BindingDB molecules. 
We report the architecture and performance of the final VAE model in \supl\ref{SI:vae}.

\textbf{Attribute Predictors.}
\label{subsec:predictors}
We train multiple property predictors for controlling generation. The architecture and performance of these predictors are reported in \supl\ref{SI:property_predictors}.
First, to test the information content of the VAE latent space, we trained multiple attribute (QED, logP, and SA) predictors on the latent embeddings.  These models show low root-mean-square-error (RMSE) on test data for all three attribute predictors.

Next, we trained a binding affinity regression model using the pIC50 ($=-log(IC50)$) data from BindingDB. This model takes a representation of a target protein sequence and latent embedding, $\rvz$, of a molecule as input, and predicts the binding affinity between the protein-molecule pair. We used pre-trained protein embeddings from ~\cite{Alley_2019} to initialize the weights for proteins. This model, along with the model for QED is used in the controlled generation pipeline. 
We also trained a binding affinity predictor using SMILES ($\rvx$)  instead of latent ($\rvz$) embedding as the input molecular representation. This model was used during the \insilico screening process as it has a higher accuracy than the model trained on the latent embeddings, comparable to a recent model described in ~\cite{karimi2018deepaffinity}.

\textbf{Selectivity Modeling.}
Selectivity to a particular target is often modeled only in the later stages of a drug development pipeline. It has been suggested that improvement in the early accounting of off-target interactions represents an opportunity to reduce safety-related attrition rates during pre-clinical and clinical development \cite{miljkovic2018data}. Given the novel nature of COVID-19, it is even more important to account for off-target selectivity in the early design state in order to minimize undesired interactions with host targets. Thus, we believe that incorporating selectivity during the candidate generation stage will contribute to a reduction in the failure rate of drug candidates. We define selectivity as the excess binding affinity (BA) of a molecule ($m$) to a target of interest ($T$) over its average binding affinities to a random selection of $k$ targets \cite{bosc2017use}: 
$Sel_{T, m} = BA(T, m) - \frac{1}{k} \sum_{i=1}^k BA(T_i, m)$.

\textbf{Controlled Generation.}
\label{controlled-gen}
Our objective is to generate molecules that simultaneously satisfy multiple (often conflicting) objectives. Specifically, we want  generated molecules controlled by high binding affinity to a selected novel SARS-CoV-2 target, high drug-likeliness, and high off-target~selectivity.   



For this purpose, we performed conditional generation using Conditional Latent (attribute) Space Sampling --- CLaSS proposed  recently in~\cite{das2020science}. In short, CLaSS leverages the attribute predictors trained on the latent features and uses a rejection sampling scheme to generate samples with desired attributes from a density model of the latent space. 
Since the goal is to sample conditionally $p(\rvx | \rva)$, where $\rva \in \R^n = [a_1, a_2, \dots, a_n]$, a set of independent attributes, CLaSS approaches this task through conditional sampling in latent space:
  $p(\rvx|\rva) = \E_\rvz[p(\rvz|\rva) p(\rvx|\rvz)]  \approx \E_\rvz[\hat{p}_\xi(\rvz|\rva) p_\theta(\rvx|\rvz)]$.
Where $\hat{p}_\xi(\rvz|\rva)$ uses rejection sampling from parametric approximations  to $p(\rvz | \rva)$. 
The term $\hat{p}_\xi(\rvz|\rva)$ is approximated using a density model $Q_\xi(\rvz)$, such as a  Gaussian mixture model and per-attribute classifier model $q_\xi(a_i|\rvz)$. This is approached by using Bayes’ rule and then conditional independence of the attributes \cite{das2020science}.  Rejection sampling is then performed through the proposal distribution: $g(\rvz) = Q_\xi(\rvz)$ that can be  directly sampled. Since we impose multiple attribute constraints for sampling, intuitively, the acceptance probability is equal to the product of the attribute predictors' scores, while sampling from  explicit density $Q_\xi(z)$. As long as there is a region in  $\rvz$ space  where $Q_\xi(\rvz) >  0$ and probabilities from all predictors are $>0$, samples will be accepted in this scheme. 
Consequently, CLaSS can sample from the targeted region of the autoencoder latent space, which was trained unsupervised. Learning to control for one or more attribute(s) in CLaSS is computationally efficient, as it does not require a surrogate model or policy learning and neither adds complicated loss terms to the original~objective. 


\textbf{\textit{In Silico} Screening.}
\label{subsec:screening}
Molecular toxicity or side effect testing is conventionally carried out via different endpoint experiments, \textit{e.g.,} \invitro molecular assays, \invivo animal testing, clinical trials, and adverse effect reports. However, these experiments are costly and time-consuming. 
We instead used a multitask deep neural network (MT-DNN) for binary (yes/no) toxicity prediction 
as an early screening tool to prioritize the testing of molecules that are less likely to be harmful and to speed up the process of finding a COVID-19 therapeutic
(For details see \supl\ref{SI:toxicity}). 
A multitask model is expected to improve the prediction by exploiting the correlation between different endpoints.
The MT-DNN was used to predict the toxicity of 12 \invitro endpoints from the Tox21 challenge \cite{Tox21}. We also predicted whether the generated molecules would fail clinical trials, using the ClinTox data~\cite{Wu2018}.

The generated molecules were screened further for target affinity and selectivity using the $\rvx$-level binding affinity predictor (See \supl\ref{SI:property_predictors}). 
To investigate the possible binding modes of the generated molecules with the target protein structure, we performed 5 independent runs of blind docking of the generated achiral molecules with the target structure using Autodock Vina \cite{trott2010autodock}. 
To evaluate the synthetic accessibility,
the generated molecules were analyzed using a retrosynthetic algorithm~\cite{schwaller2020predicting} based on the Molecular Transformer~\cite{SchwallerFWD} trained on patent chemical reaction data.

\section{Results and Discussion}

\subsection{Benchmarking Molecular VAE Model}
The architecture and performance metrics of the final VAE model that is adaptively pre-trained from ZINC to BindingDB with SA and QED supervision are provided in \supl\ref{SI:vae}, along with a comparison to a number of baseline models. The majority of the generated molecules are chemically valid (90\%), unique (99\%), pass relevant filters (95\%), and show a slightly higher  diversity (Table \ref{Unsup1}). One interesting observation from Table \ref{Unsup2} is that the generated molecular ensemble  has a higher Fréchet ChemNet Distance (FCD) \cite{preuer2018frechet} with respect to chemical scaffolds present in both ZINC and BindingDB training molecules. This implies that adaptive pre-training from ZINC to BindingDB enables the discovery of novel chemical scaffolds, which is further confirmed by comparing the Tanimoto Similarity between generated scaffolds and reference scaffolds (\supl Figure \ref{fig:nov-scaffolds-hist}). A few synthetically plausible and novel scaffolds from the generated set are shown in \supl Figure~\ref{fig:nov-scaffolds}

\subsection{Attributes of COVID-Targeted Molecules}

\begin{figure}[!b]
  \centering
  \begin{subfigure}[b]{0.31\textwidth}
    \centering
    \includegraphics[width=\textwidth]{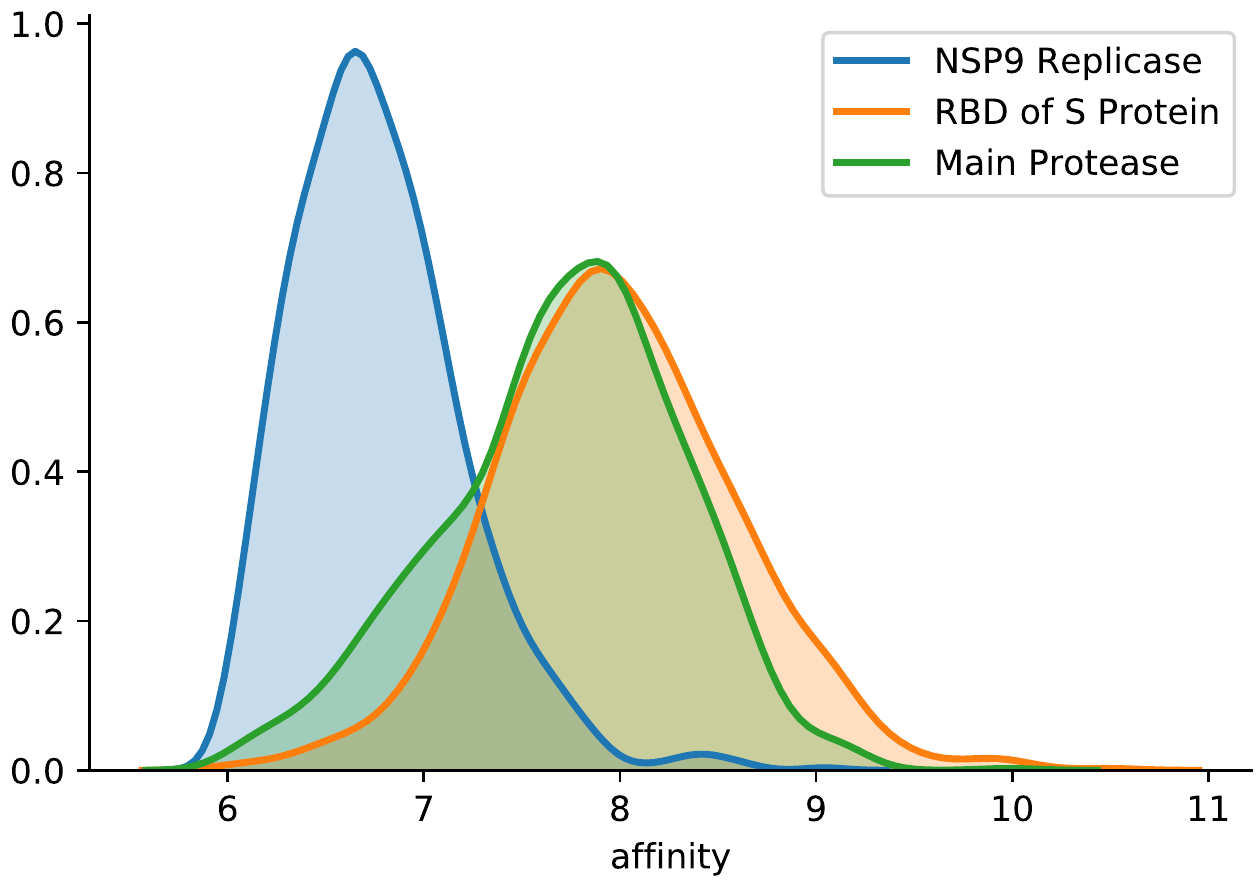}
    \label{fig:density_AFF}
  \end{subfigure}
  \hfill
  \begin{subfigure}[b]{0.367\textwidth}  
    \centering 
    \includegraphics[width=\textwidth]{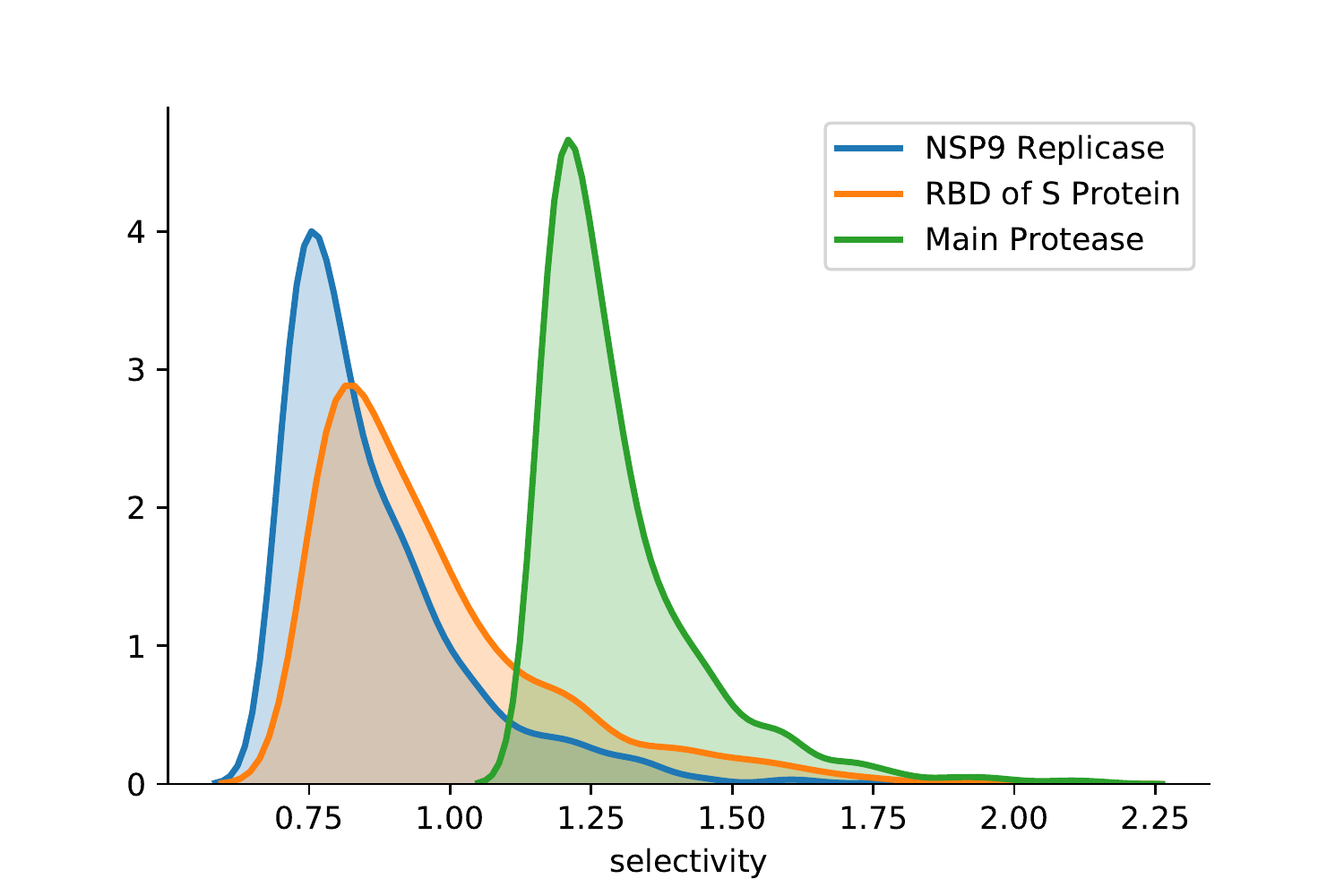}
    \label{fig:density_SEL}
  \end{subfigure}
  \hfill
  \begin{subfigure}[b]{0.303\textwidth}   
    \centering 
    \includegraphics[width=\textwidth]{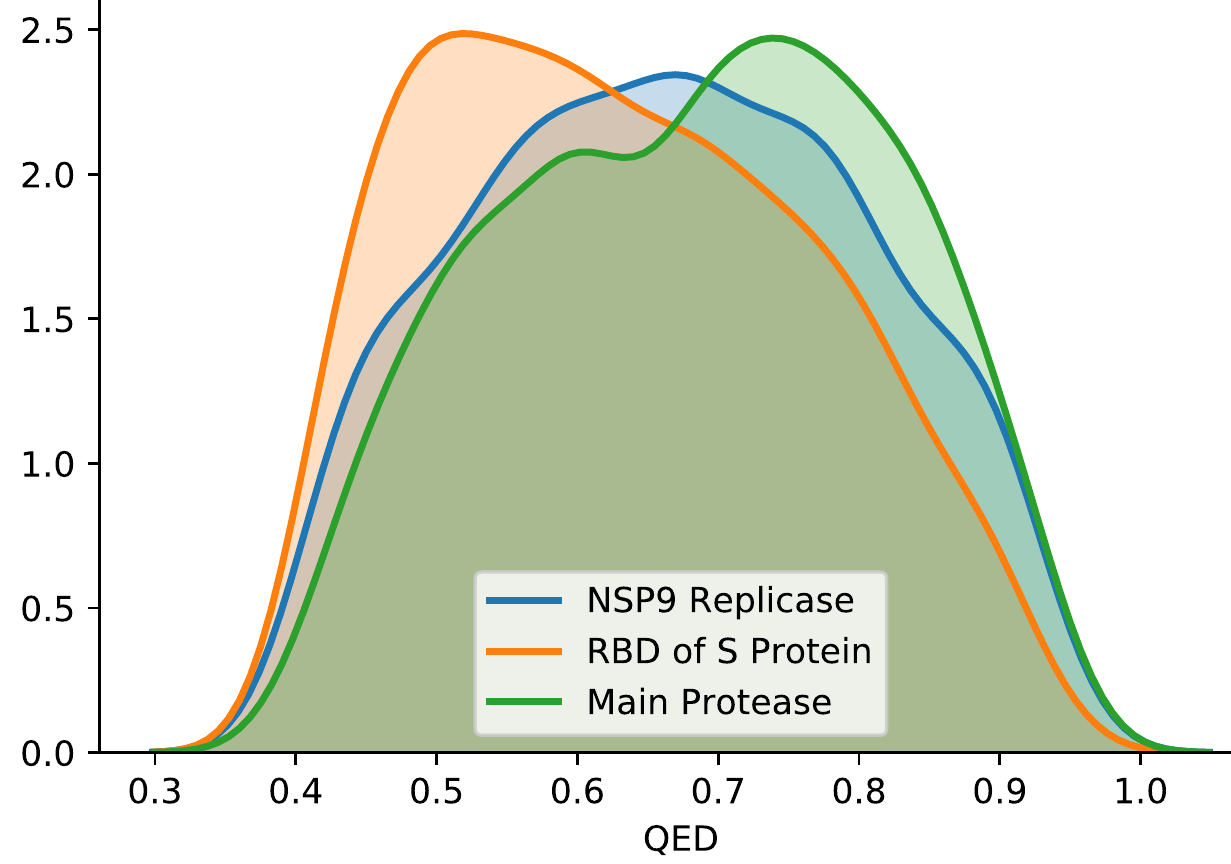}
    \label{fig:density_QED}
  \end{subfigure}
  \caption{Density plots of (a) binding affinity, (b) off-target selectivity, (c) QED for selected molecules.}%
  \label{fig:density}%
\end{figure}

\textbf{CogMol-Controlled Attributes --- Target Affinity, Selectivity, and QED.}
Table \ref{Class_prob} reports  higher  proportion of molecules with desired attributes in the set accepted in CLaSS, when compared to a randomly sampled set, implying that CLaSS does generate a more optimal set than random sampling from the latent space, and the success depends on the target context.
We further selected around 1000 CogMol-generated molecules for each target as explained in \supl\ref{SI:selected_set}. 
The 
density plots in Figure~\ref{fig:density} of the selected set 
indicate that generating high-affinity ligands is more challenging for  NSP9  (Figure~\ref{fig:density}a), while M\textsuperscript{pro}  ligands are more selective in general (Figure~\ref{fig:density}b), which is likely due to relative novelty of the target sequences with respect to BindingDB training sequences  (see \supl\ref{SI:targets}). The QED distribution also highlights target sequence dependence of the generated molecules, as the molecules targeting RBD show a peak at a lower QED value in the distribution. Several randomly chosen samples 
for each SARS-CoV-2 target are shown in \supl Figure~\ref{fig:generated_molecules}.
 
\begin{table}[!tb]
\centering
\caption{Normalized fraction of molecules that are accepted in CLaSS with different set of controls (Affinity, QED, and Selectivity). The values of controls are normalized between 0 and 1. As we increase the extent of controls, a relatively higher proportion of molecules meeting all criteria are in the accepted set compared to a randomly sampled set.}
\label{Class_prob}
\scalebox{0.85}{
\begin{tabular}{
        r|
        S[table-format=1.3]
        S[table-format=1.3]
        S[table-format=1.3]
        S[table-format=1.3]
        S[table-format=1.3]
        S[table-format=1.3]
        S[table-format=1.3]
        }
    \toprule
    & \multicolumn{2}{c}{Aff \textgreater 0.5} & \multicolumn{2}{c}{Aff \textgreater 0.5 \& QED \textgreater 0.8} & \multicolumn{3}{c}{Aff \textgreater 0.5 \& QED \textgreater 0.8 \& Sel \textgreater 0.5} 
    \\
    & {Accepted} & {Random} & {Accepted} & {Random} & & {Accepted} & {Random}
    \\
    \midrule
    NSP9 & 0.567 & 0.355 & 0.45 & 0.211 && 0.069 & 0.007     \\ 
    RBD  & 0.546 & 0.369 & 0.429 & 0.217 && 0.09 & 0.009 
    \\
    M\textsuperscript{pro} & 0.603 & 0.366 & 0.472 & 0.216 && 0.104 & 0.011 
    \\
    \bottomrule                
\end{tabular}
}
\end{table}

\textbf{Novelty.}
The novelty distributions, as estimated using the Tanimoto Similarity \cite{tanimoto_willett} between molecular fingerprints, of the generated molecules with respect to both the PubChem~\cite{pubchem_kim} database and our training set are shown in \supl Figures~\ref{fig:density_NOV_train} and ~\ref{fig:density_NOV_pubchem}. 
When compared with the training database of size $\sim$ 1.9 M, we find that the likelihood of generating molecules with a novelty value of 0 is $\leq$ 2\%.  With respect to the larger PubChem database consisting of $\sim$ 103 M molecules, the majority of which were not included in model training, we find the percentage of generated molecules with novelty value of 0 is 9.5\%, 3.7\%, and 8.3\%  generated molecules for M\textsuperscript{pro}, RBD, and NSP9, respectively. Higher FCD of those generated molecules with respect to test scaffolds in ZINC/BindingDB  (\supl Table \ref{SI:FCD}) further confirms presence of novel chemical scaffolds in them.

\begin{table}[!bt]
\centering
\caption{CogMol-generated SMILES found in PubChem and their predicted  affinity (pIC50), lowest docking free energy (kcal/mol), PubChem Compound ID (CID), and reported biological activity. }
\label{fig:pubchem-match}
\scalebox{0.85}{
\begin{tabular}{r|
                S[table-format=1.2]
                S[table-format=1.1]
                r
                l}
\toprule
{Target} & {Pred.~Affinity} & {Docking Energy} & {CID} & {Biological Activity} \\
\midrule
\multirow{3}{*}{\shortstack[r]{NSP9\\Dimer}} & 6.51 & -7.7 & 12042753 & Antagonist of rat mGluR\\
& 7.06 & -5.6 & 44397285 & Active to human S6 kinase\\
& 7.18 & -6.4 & 10570770 & Matrix metalloproteinase inhibitor\\[6pt]

\multirow{2}{*}{\shortstack[r]{Main\\Protease}} & 7.24 & -6.1 & 10608757 & Dihydrofolate reductase inhibitor\\
& 6.91 & -6.9 & 872399 & Shiga toxin inhibitor \\
[6pt]

\multirow{1}{*}{RBD } & 7.82 & -7.5 & 76332092 & Plasmepsin inhibitor \\
\bottomrule
\end{tabular}}
\vspace{-5pt}
\end{table}

\textbf{CogMol Identifies PubChem Molecules with Potential Anti-COVID Activity.} 
Only 19, 5, and 15 of the generated molecules match exactly with an existing SMILES string in PubChem, for M\textsuperscript{pro}, RBD, and NSP9, respectively. 
Some of these SMILES are reported with \textit{biological activity} in PubChem, as shown in Table \ref{fig:pubchem-match},  which calls for further investigation.  For example, the molecule with PubChem Compound ID (CID) 76332092 
is a known Plasmepsin-2 and Plasmepsin-4 inhibitor and 
has also shown antimalarial activity against chloroquine-sensitive Plasmodium falciparum.
As RBD from S protein binding to angiotensin-converting enzyme-2 (ACE-2) receptor is needed for the viral entry to the  host cells \cite{hoffmann2020sars}, both RBD and ACE-2 receptor are being actively investigated as COVID-19 targets.   Chloroquine (and its hydroxy derivative) is a known ACE-2 inhibitor and has been already considered as a promising COVID-19 drug \cite{liu2020hydroxychloroquine}. 
CID 76332092 deserves further investigation in the context of SARS-CoV-2 as it shows a predicted pIC50 of 7.82 and lowest docking binding free energy (BFE) of -6.80 kcal/mol (Figure~\ref{fig:rbd_GEN146_docking}) to the ACE-2 binding pocket of RBD (Table \ref{fig:pubchem-match}).
The generated molecule with highest predicted affinity for RBD (with a pIC50 of 10.49 and a docking BFE of -6.9 kcal/mol in the top binding mode with ACE-2 binding pocket) also shares a strong maximum common subgraph similarity \cite{cao2008maximum} with Telavancin, an approved skin infection and Pneumonia drug, as shown in Figure~\ref{fig:telavancin}. 
These results indicate that  CogMol can  generate promising and biologically relevant drug candidates beyond the training dataset.

\textbf{Docking with Target Structure.}
\label{subsec:docking}
Table \ref{tbl:docking_1K} summarizes these results. In the best (lowest BFE) docking pose, 87\%, 91\%, and 95\% of generated molecules show a minimum BFE of $<-6$ kcal/mol for NSP9 dimer, M\textsuperscript{pro}, and RBD, respectively. For each target, we classified each molecule by its binding location, fitting the geometric centers of docked molecules drawn from a larger set of 875K samples 
to a mixture of 4, 5, and 6 Gaussian models, respectively (see \supl\ref{app:docking_analysis}). 
We also report the average and minimum BFE, as well as the fraction of generated molecules with a BFE of $<-6$ kcal/mol for each cluster (Table \ref{tbl:docking_1K}). Results show that even though CLaSS used only target sequence information for controlled generation, generated molecules do identify the relevant and known druggable binding pockets within the 3D target structure and  bind to those favorably.  


\begin{table}[!tb]
\centering
\caption{Docking analysis: Size, average ($E$) ($\pm$ standard deviation) binding free energy (BFE), minimum BFE, fraction of generated molecules with BFE $<-6$ kcal/mol for each cluster. In parentheses after target name:\% of  generated molecules for the respective target that had a BFE $<-6$ kcal/mol. Only top 2 clusters are shown (see \supl\ref{tbl:docking_1K_full}).}
\label{tbl:docking_1K}
\scalebox{0.85}{
\begin{tabular}{rl|
                S[table-number-alignment=center]
                S[table-number-alignment=center,separate-uncertainty,table-figures-uncertainty=1,table-figures-decimal=1]
                S[table-number-alignment=center,table-format=1.1]
                S[table-number-alignment=center]}
\toprule
\multicolumn{2}{c|}{Target} & {Size (\%)} & {$E$ (kcal/mol)} & {Min (kcal/mol)} & {Low Energy (\%)} \\
\midrule
\multirow{2}{*}{NSP9 Dimer (87\%)} & cluster 0 & 67 & -6.8 \pm 0.7 & -8.6 & 88\\
& cluster 1 & 22 & -6.9 \pm 0.9 & -8.8 & 85\\
[6pt]

\multirow{2}{*}{Main Protease (91\%)} & cluster 0 & 76 & -7.2 \pm 0.8 & -9.5 & 93\\
& cluster 1 & 18 & -6.9 \pm 0.8 & -9.2 & 86\\
[6pt]

\multirow{2}{*}{RBD (95\%)} & cluster 0 & 30 & -6.9 \pm 0.6 & -8.3 & 93\\
& cluster 1 & 36 & -7.2 \pm 0.6 & -9.1 & 97\\
\bottomrule
\end{tabular}
}
\vspace{-5pt}
\end{table}



\subsection{CogMol-generated Molecules Targeting Human HDAC1}
\label{sec:hdac}
Human HDAC1  plays key role in eukaryotic gene expression and is  implicated in cancer. Though it is present in BindingDB, there are only  a handful of molecules with high QED  and high pIC50, see Table \ref{tab:hdac1}.  We applied CogMol to generate optimal ligands targeting HDAC1. Table \ref{tab:hdac1} shows that CogMol-generated molecules comprise a larger proportion of molecules satisfying high pIC50 and QED criteria, implying CogMol can discover novel and optimal molecules even in a low-data regime.  

\subsection{Synthesizability and Toxicity of Generated Molecules}
\label{subsec:synthesis}

The number of steps/reactions needed to complete the synthesis (synthetic design) provides an estimate of the complexity of the molecules respect to commercially available materials (more details and the parameters adopted, can be found in the \supl\ref{SI:retrosynthesis}).
\begin{figure}[t!]
  \centering
  \hfill
  \begin{subfigure}[b]{0.45\textwidth}
    \includegraphics[width=\textwidth]{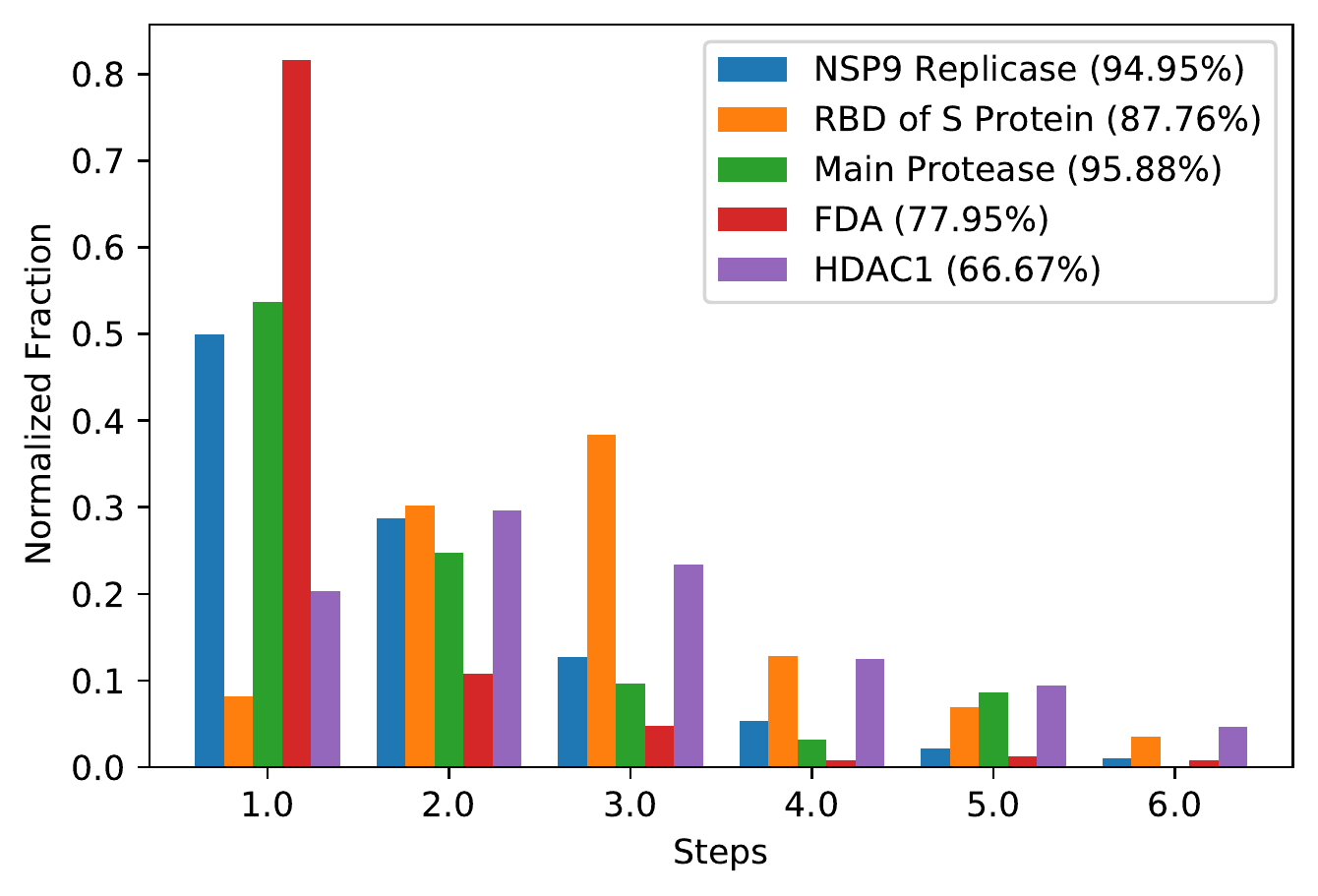}
  \end{subfigure}
  \hfill
  \begin{subfigure}[b]{0.45\textwidth}
    \includegraphics[width=\textwidth]{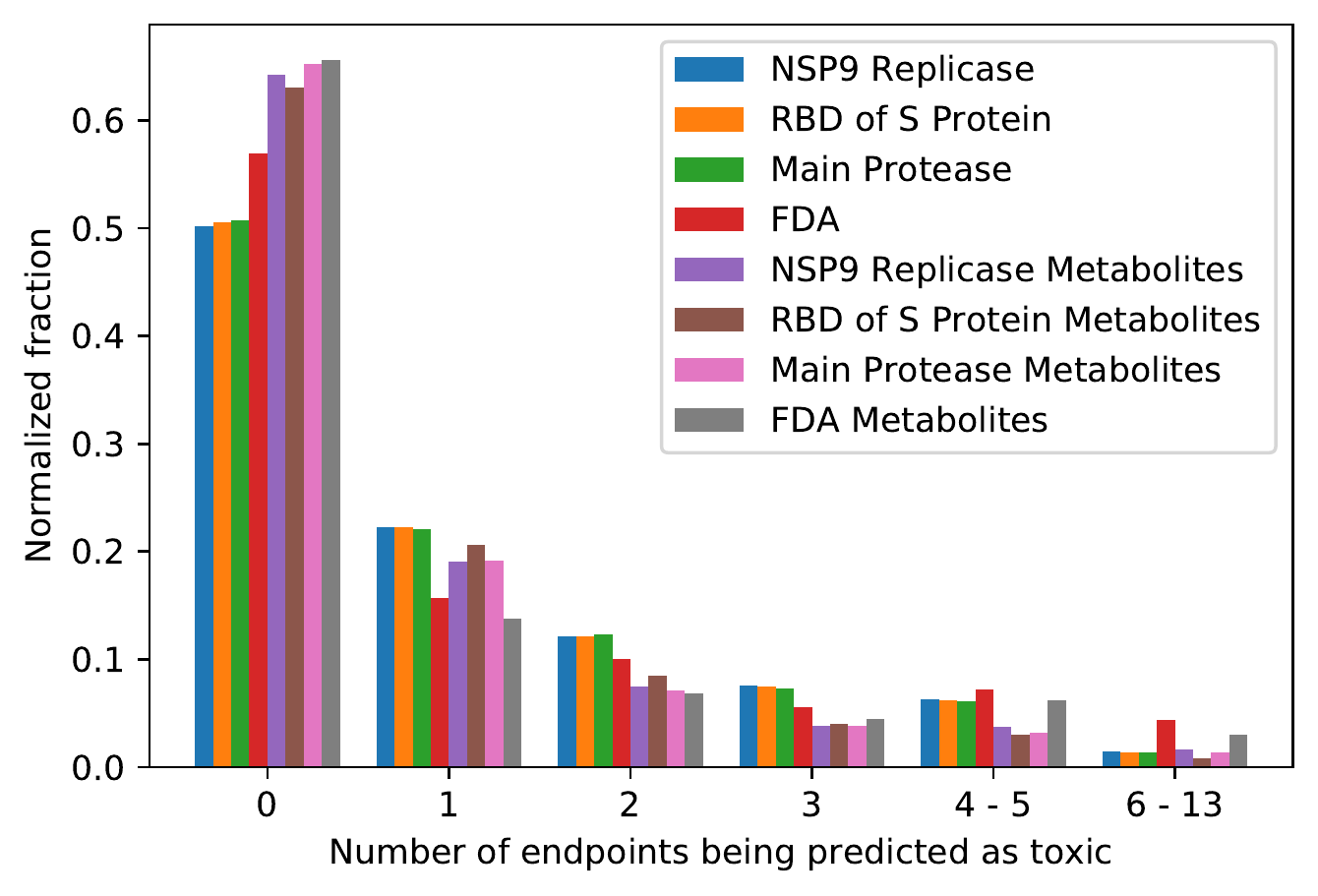}
  \end{subfigure}
  \hfill
  \hfill
  \caption{
  (a) 
  Bar plots describe the percentage of molecules synthesizable for the exact number of retrosysthesis steps. 
  Legend reports the fraction of molecules for each set marked as synthetically accessible. (b) Percentage of parent molecules or their metabolites as a function of number of endpoints in which predicted to be toxic.
  }
  \label{fig:synthesis_toxicity}
  \vspace{-5pt}
\end{figure}
In Figure~\ref{fig:synthesis_toxicity}a (legend), we report the percentage of feasibility  for 4 sets of generated molecules, each targeting a different protein -   NSP9, RBD, M\textsuperscript{pro}, and HDAC1. For SARS-CoV-2 targets, molecules were selected by considering the top-100 molecules based on SA. For  HDAC1, calculation was done on a set of 100 generated molecules.
We also estimated feasibility for a selection of FDA-approved and commercially available drugs~\cite{FDAenamine} (FDA), used as a baseline. 
For the FDA set, the fraction of molecules predicted as feasible is $\sim$ 78\%.
This is expected, since most of these molecules have been protected with patents and are therefore chemically accessible with the reaction knowledge available in patents.
The molecules generated by CogMol for the three COVID-19 targets perform better than the FDA with successful rates $>$ 85-90\%. The HDAC1 set instead shows a value of $\sim$ 67\%.
The higher success rate for the COVID target sets indicates that the molecules in these classes are easier to synthesize from commercially available materials than the molecules belonging to the FDA class.
The HDAC1 set, while  still showing a relatively high synthesizability rate, demonstrates additional need 
of chemical knowledge uncovered in patents.
The  distribution of the number of steps needed for each set reveals two interesting observations, as  shown in Figure~\ref{fig:synthesis_toxicity}a: FDA, M\textsuperscript{pro} and NSP9 molecules show a peak at 1 step and HDAC1 and RBD showing a peak respectively at 2 and 3 steps.
In comparison, $>$ 80\% of the successfully synthesizable FDA approved molecules can be made in a single step from commercially available molecules, likely because their precursors are also made commercially available after approval.
The M\textsuperscript{pro} and NSP9 sets are similar to the FDA approved drugs. They are characterized by a lower degree of complexity, indicating close relation to commercially available molecules,  
when compared to HDAC1 and RBD.
Overall, the retrosynthetic analysis of the generative model outcomes clearly shows that the generated structures are chemically relevant and synthetically feasible.
Additional results revealing correlations between number of  synthesis steps and  properties of molecules for the COVID-19 targets can be found in \supl\ref{SI:retrosynthesis}.

Figure \ref{fig:synthesis_toxicity}b shows toxicity analyses of the CogMol-generated molecules, their metabolites, as well as of FDA-approved drugs by using the MT-DNN model.  A molecule was considered toxic if it was predicted to be toxic in $\ge$ 2 endpoints.
The metabolites were predicted by using a recently proposed work that models the human metabolite prediction task of small molecules as a sequence translation problem and uses a Seq2Seq Transformer model originally pre-trained on chemical reaction data and further fine-tuned  on metabolite reaction data to predict the outcome of human metabolic reactions \cite{litsa_das_kavraki_2020}. Results in Figure \ref{fig:synthesis_toxicity}b show that majority ($\sim$ 70\%) of the generated molecules, as well as their predicted metabolites ($\sim$ 80\% of them) are predicted toxic only in 0-1 endpoints out of a total of 13, which is comparable to the FDA-approved drugs. 

\subsection{Sharing and Visualization of Generated Molecules}
We share around $\sim$ 3500 generated molecules under an open license for the research community to download and evaluate. In order to help domain experts, we also created a publicly available Molecule Explorer tool to facilitate screening and filtering of the molecules, perform novelty analysis, and identify closest molecules in PubChem.  
A screen cast of the Molecule Explorer tool is provided.\footnote{\url{https://www.youtube.com/watch?v=cYb8\_catBpI}} 

\section{Conclusions and Future Work}
In this paper, we proposed CogMol, a framework for Controlled  Generation of Molecules with a set of desired attributes. Our framework can handle targeted and novel compound generation for multiple proteins using the same trained model, can generalize to unseen viral proteins, and accounts explicitly for off-target selectivity. 
Additionally, we provide an \insilico screening method that accounts for target structure binding, \textit{in vitro} and clinical toxicity of parent molecules and their metabolites, and synthesis feasibility.
When applied to COVID-19 novel viral protein sequences, CogMol generated novel molecules that were able to bind favorably to the relevant druggable  pockets of the target structure. 
The generated compounds are also comparable to FDA-approved drugs in parent molecule, metabolite toxicity, and synthetic feasibility.
In summary, our framework provides an efficient and viable computational framework for de novo multi-objective design and filtering of optimal drug compounds that are specific and selective to novel/unseen targets. 
Future work will address accounting for additional contexts (on top of target protein), adding  other pharmacologically relevant controls, and also weigh those according to  their relative importance to make CogMol framework more efficient in term of generating promising drug candidates. 
 
\section{Statement of Broader Impact}
We discuss the broader impact of our work from the following perspectives.

\textbf{Benefits}
To date, SARS-CoV-2 has infected millions and killed hundreds of thousands around the globe and continues to cause a severe economical crisis~\cite{fernandes2020economic}.
No approved therapeutics is available against any coronavirus~\cite{agostini2018coronavirus}, no SARS-CoV-2 vaccine has successfully completed phase II human trails~\cite{le2020covid}, and drugs for repurposing 
are still undergoing investigation~\cite{wang2020remdesivir}. Therefore, it is timely  to explore for efficient \textit{de novo} drug design approaches to combat COVID-19 and future pandemics.
The CogMol framework is adapative, generic, and could pave the road for accelerated discovery of new antivirals optimized against specific SARS-CoV-2 (or other novel virus) targets. This could have a major impact on our global effort against COVID-19 and future novel pandemics and save human lives.   

We demonstrated that our framework can generate  target-specific and selective compounds for \textit{unseen} protein targets, a novel property that may be key for swift reactions to possible SARS-CoV-2 mutants. We further provide  \textit{early assessment} of novel AI-generated compounds on target structure binding, and synthetic feasibility and toxicity in the context of FDA-approved drugs, in order to identify a list of promising compounds that is of reasonable size  and can be immediately sent to wet lab for synthesis and validation.  We showed the efficiency of the  framework in terms of handling multiple constraints at once and can be easily extended to   adding more controls to account for additional factors considered crucial in  drug discovery such as ADME properties. Thus, our approach systematically bridges biology and machine learning to accelerate drug discovery.

We further share with the community a list of CogMol-generated compounds (and their attributes) designed for three novel SARS-Cov-2 targets, as well as a molecular explorer tool to visualize,  experience, and provide feedback on these molecules. This sets our vision for an open community of discovery that facilitates interactions between AI researchers and medicinal scientists.

\textbf{Risks and the Potential to Cause Harm}
While our approach offers enormous potential to speed up the development of new drugs, it must be realized that drug candidate generation and \insilico screening are merely first steps in the development of viable therapeutics. The ability of the public to order these novel compounds online, poses a risk that it might be tried by people who are not sufficiently educated about the dangers of exposing themselves to these molecules in an uncontrolled setting. The public must be educated to not to treat these candidates as approved drugs or miracle cures. Further, since our framework allows generation of molecules satisfying arbitrary objectives, this capability can be misused by bad actors to design potentially harmful chemicals. 

\textbf{Consequences of Failure}
It is possible that our framework will not be able to generate molecules with desired properties either because of bias in training data or because of the inaccuracy of the predictors used for controlled generation. In this case, the properties of these molecules should be independently validated by using multiple independent mechanisms.

\clearpage

\bibliographystyle{ieeetr}
\bibliography{confs.bib}

\clearpage
\appendix
\counterwithin{figure}{section}
\counterwithin{table}{section}
\section*{Supplementary Materials}

\section{Protein Targets Chosen for Generation}
\label{SI:targets}
\begin{figure*}[ht]
    \centering
    \includegraphics[width=0.9\textwidth]{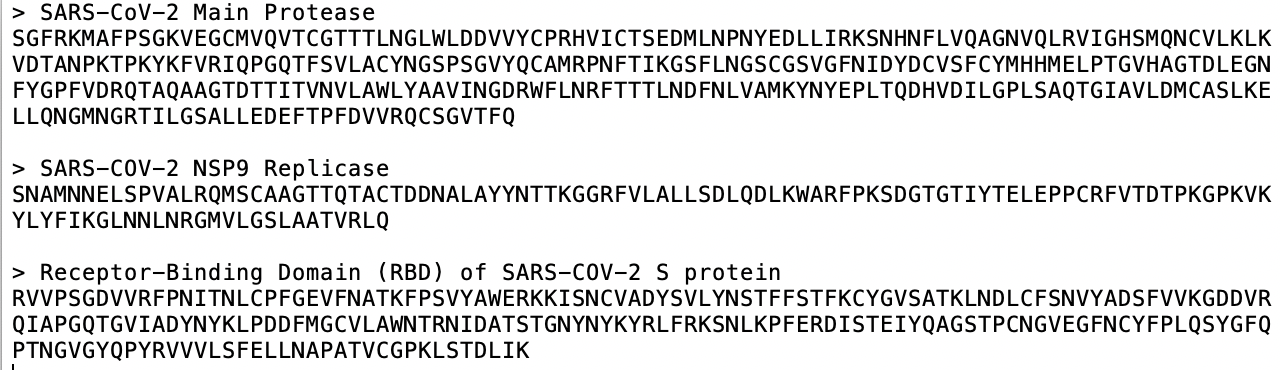}
    \caption{Sequences of SARS-CoV-2 targets}
    \label{fig:targets}
\end{figure*}

Figure~\ref{fig:targets} shows the amino acid sequences corresponding to the three SARS-CoV-2 targets. 

We also computed the similarity of these targets with respect to our training data from BindingDB using NCBI-BLAST tool\footnote{\url{https://ftp.ncbi.nlm.nih.gov/blast/executables/blast+/LATEST/}}.
The BLAST tools computes Expect value (E-value) - a measure of statistical  significance of the match between the query sequence and database sequences. Larger E-value indicates a higher chance that the similarity between the hit (from the database) and the query is merely a coincidence, i.e. the query is not homologous or related to the hit. 
The lowest Expect value with respect to BindingDB protein sequences using the default parameters of BLAST are: M\textsuperscript{pro} = 0.51 (query coverage = 40\%), RBD = 1.9 (query coverage =  26\%), NSP9 = 3.2 (query coverage = 10\%).


\FloatBarrier

\section{Variational Autoencoder}
\label{SI:vae}
We used a  bidirectional Gated Recurrent Unit (GRU) with a linear output layer as an encoder. The decoder is a 3-layer GRU RNN of 512 hidden dimensions with intermediate dropout layers with dropout probability 0.2. The performance characteristics of the VAE model with respect to various metrics are given below.

\begin{table}[!ht]
\centering
\caption{The performance metrics of the generative model: Fraction of valid molecules, Fraction of unique molecules from a sample of 1,000 and 10,000 molecules, Internal Diversity (IntDiv1 and IntDiv2), Fraction of molecules passing filters (MCF, PAINS, ring sizes, charges, atom types)}
\label{Unsup1}
\scalebox{0.85}{
\begin{tabular}{
    r|
    S[table-format=1.4]
    S[table-format=1.1]
    S[table-format=1.4]
    S[table-format=1.4]
    S[table-format=1.4]
    S[table-format=1.4]
}
\toprule
Model & {Valid} & {Unique@1k} & {Unique@10k} & {IntDiv1} & {IntDiv2} & {Filters}  \\
\midrule
ZINC (Unsupervised) & 0.9553 & 1.0 & 0.9996 & 0.8568 & 0.8510 & 0.9889 \\
ZINC (Supervised) & 0.95 & 1.0 & 0.999 & 0.8578 & 0.8521 & 0.9888  \\
BindingDB (Supervised) & 0.904 & 1.0 & 0.9993 & 0.8717 & 0.8665 & 0.9482  \\
CharRNN & 0.809 & 1.0 & 1.0 & 0.855 & 0.849 &  0.975 \\
AAE & 0.997 & 1.0 & 0.995  & 0.857 & 0.85  &  0.997 \\
VAE & 0.969 & 1.0 & 0.999 & 0.856 & 0.851 &  0.996  \\
JT-VAE & 1.0 & 1.0 & 0.999 & 0.851 & 0.845 & 0.978  \\
Training & 1.0 & 1.0 & 1.0 & 0.857 & 0.851 & 1.0  \\
\bottomrule
\end{tabular}}
\end{table}

\begin{table}[!htb]
  \centering 
  \caption{Performance evaluation of the generative model using scaffold split metrics: Fréchet ChemNet Distance (FCD), Similarity to the nearest neighbour (SNN), Fragment similarity (Frag), Scaffold similarity (Scaff). The model trained on BindingDB is evaluated on both BindingDB and ZINC Scaffolds.}
  \label{Unsup2}
  \scalebox{0.85}{
  \begin{tabular}{
    r|
    S[table-format=1.4]
    S[table-format=1.4]
    S[table-format=1.4]
    S[table-format=1.4]
    S[table-format=1.4]
    S[table-format=1.4]
    S[table-format=1.4]
    S[table-format=1.4]
  }
    \toprule
    \multicolumn{1}{r}{\multirow{2}{*}{Model}} &
      \multicolumn{2}{c}{FCD} &
      \multicolumn{2}{c}{SNN} &
      \multicolumn{2}{c}{Frag} &
      \multicolumn{2}{c}{Scaff} \\
    & {Test} & {TestSF} & {Test} & {TestSF} & {Test} & {TestSF} & {Test} & {TestSF} \\
    \midrule
    ZINC(Unsupervised) & 0.166 & 0.603 & 0.560 & 0.533 & 0.999 & 0.997 & 0.905 & 0.128\\
    ZINC(Supervised) & 0.2051 & 0.6222 & 0.5526 & 0.5267 & 0.999 & 0.998 & 0.8907 & 0.1319\\
    BindingDB (BindingDB Scaff.) & 0.7322 & 9.535 & 0.4335 & 0.3732 & 0.998 & 0.8493 & 0.5382 & 0.0764\\
    BindingDB (ZINC Scaff.) & 7.3416 & 7.7179 & 0.4089 & 0.4002 & 0.9593 & 0.9576 & 0.3196 & 0.0869\\
    CharRNN & 0.355 & 0.899 & 0.536 & 0.514 & 0.999 & 0.996 & 0.882 & 0.14 \\
    AAE & 0.395 & 1.0 & 0.62 & 0.575 & 0.995 & 0.994 & 0.866 & 0.1 \\
    VAE & 0.084 & 0.541 &  0.623 & 0.677 & 1.0 & 0.998 & 0.993 & 0.062\\
    JT-VAE & 0.422 & 0.996 & 0.556 & 0.527 & 0.996 & 0.995 & 0.892 & 0.1\\
    Training & 0.008 & 0.476 & 0.642 & 0.586 & 1.0 & 0.999 & 0.991 & 0.0 \\
    \bottomrule
  \end{tabular}}
\end{table}

\begin{figure}[htb]
    \centering
    \includegraphics[width=0.82\textwidth]{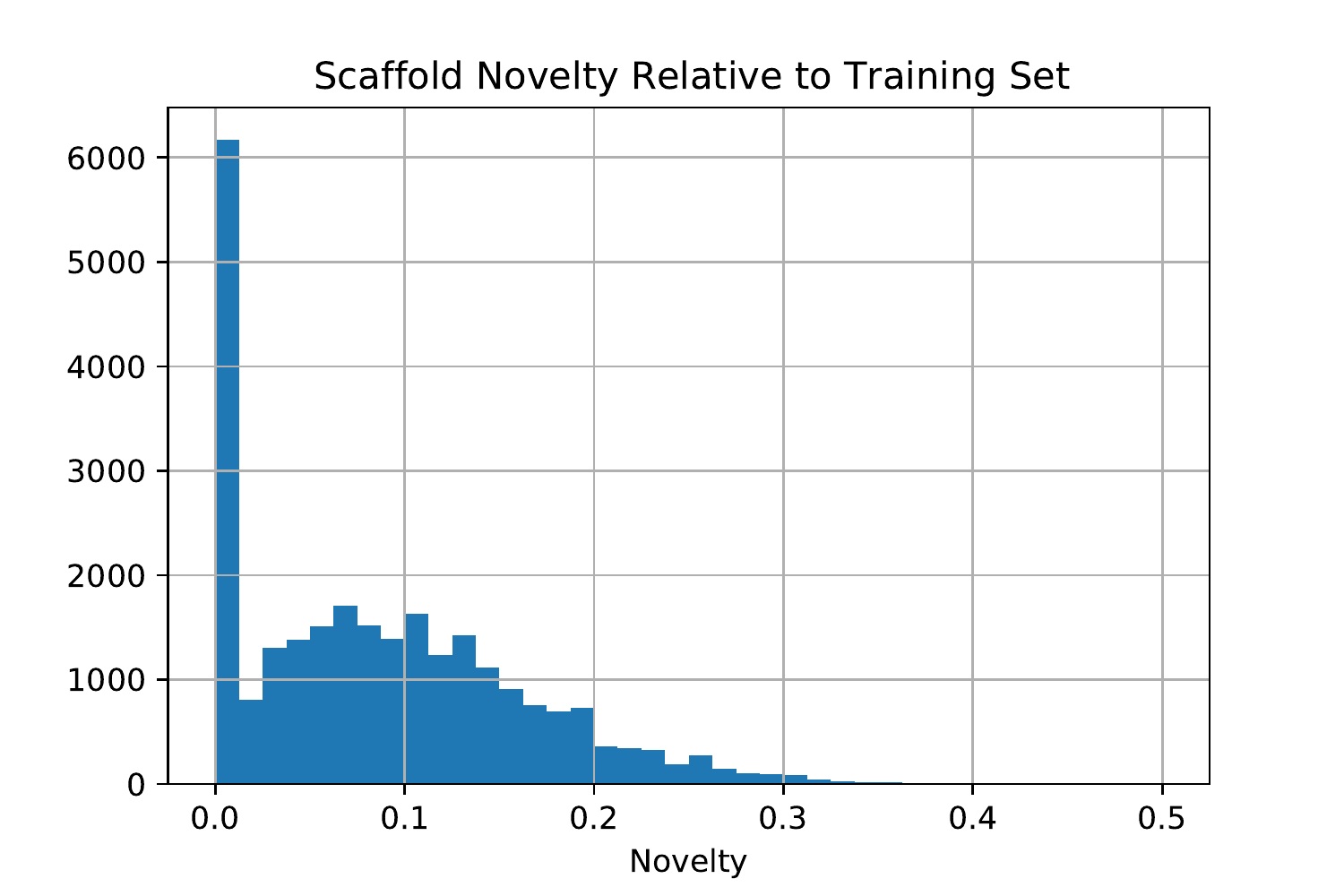}
\caption{The novelty of the scaffold of each generated molecule compared to the most similar scaffold in the training set. 26k molecules were generated and their scaffolds compared to the scaffolds of every molecule in the Zinc and BindingDB datasets. The results indicate that while many molecules have scaffolds which are present in the dataset (indicated by the spike at novelty = 0), there are many molecules that contain scaffolds not at all present in the training data.}
  \label{fig:nov-scaffolds-hist}
\end{figure}

\begin{figure}[ht]
    \includegraphics[width=\textwidth]{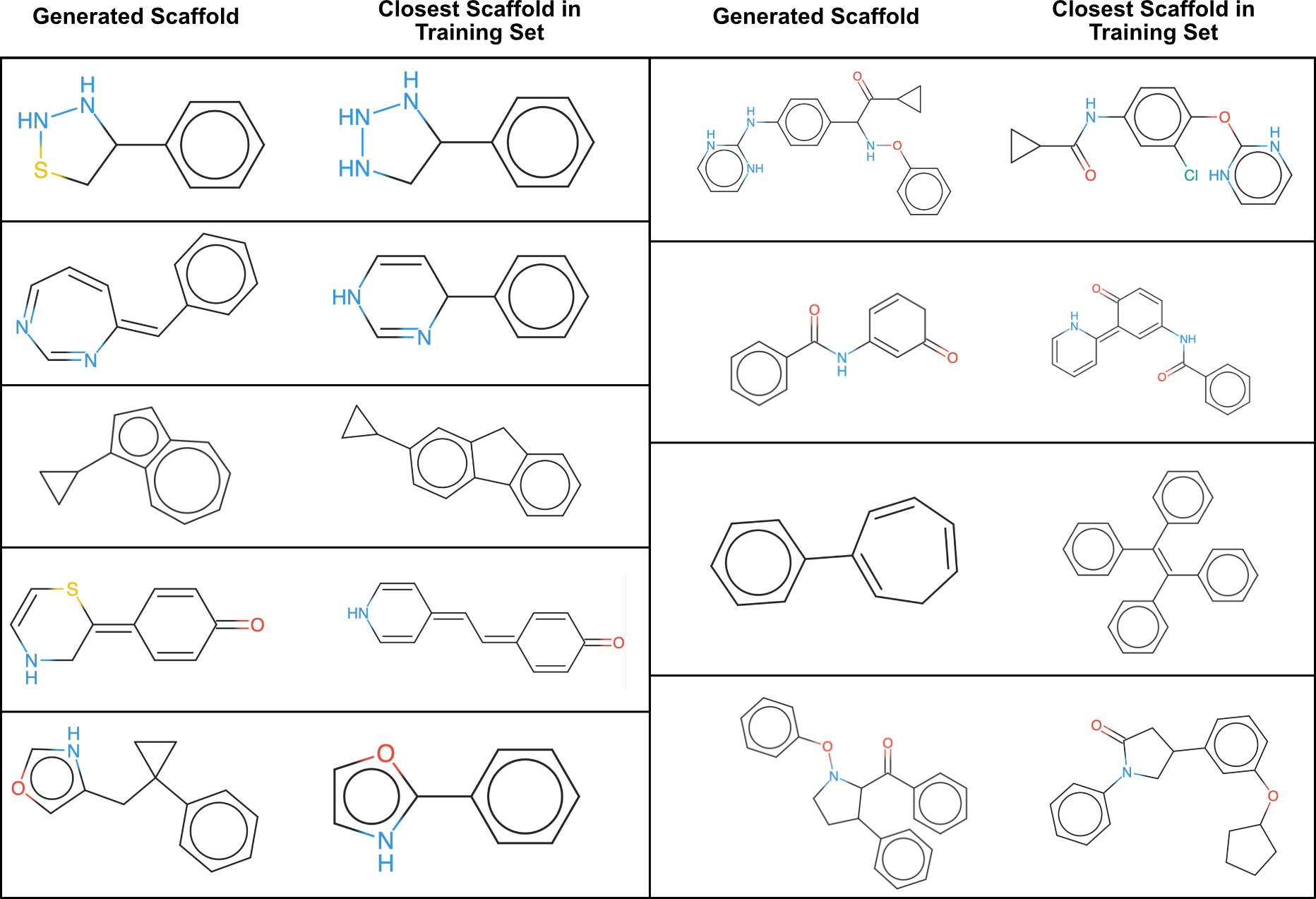}

\caption{Comparing example scaffolds in the 26k generated molecules to the scaffolds within the training data. The most similar scaffold in the training dataset, as calculated by the Tanimoto Similarity of the fingerprints, is shown next to the scaffold of each generated molecule. These novel scaffolds were determined by a synthetic organic chemist to be synthetically plausible.}
  \label{fig:nov-scaffolds}
\end{figure}

\FloatBarrier

\section{Selection of Molecules for Further Analysis}
\label{SI:selected_set}

We selected around 1000 molecules for each target based on binding affinity ($pIC50 >6$), QED ($> 0.4$), Synthetic Accessibility ($<5$), number of toxic endpoints ($< 2$), logP ($ < 5$) and Mol. Wt ($ < 500 $) for further analysis. The off-target selectivity was chosen to be higher than $1.15$, $0.75$ and $0.7$ for Main Protease, RBD of S Protein and NSP9 Replicase respectively. 


\FloatBarrier
\section{Random Examples of Generated Molecules}
\begin{figure*}[htb]%
    \centering
    \hfill
    \includegraphics[width=\textwidth]{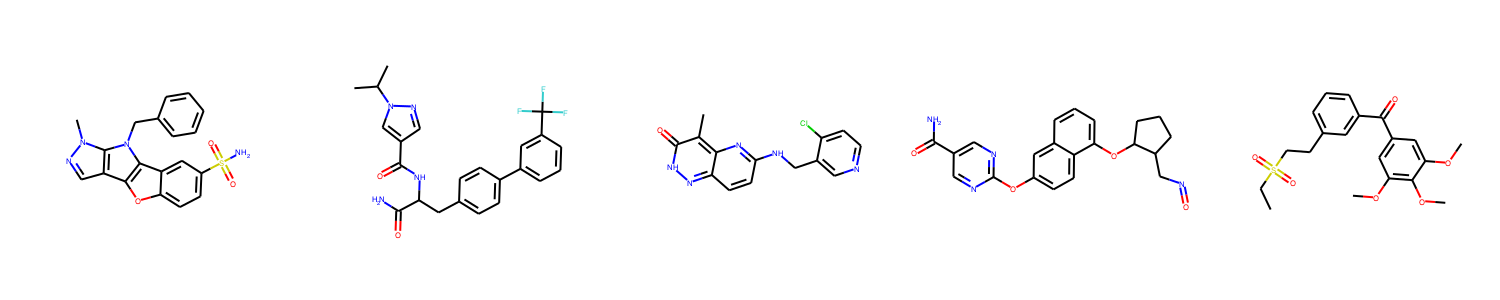}  
    \hfill
    \rule{0.92\textwidth}{0.3pt}  
    \hfill
    \includegraphics[width=\textwidth]{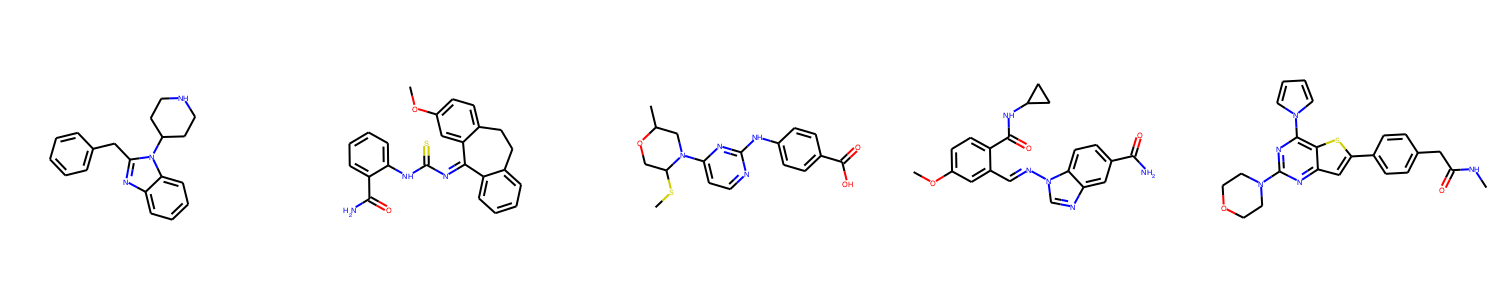}  
    \hfill
    \rule{0.92\textwidth}{0.3pt}  
    \hfill
    \includegraphics[width=\textwidth]{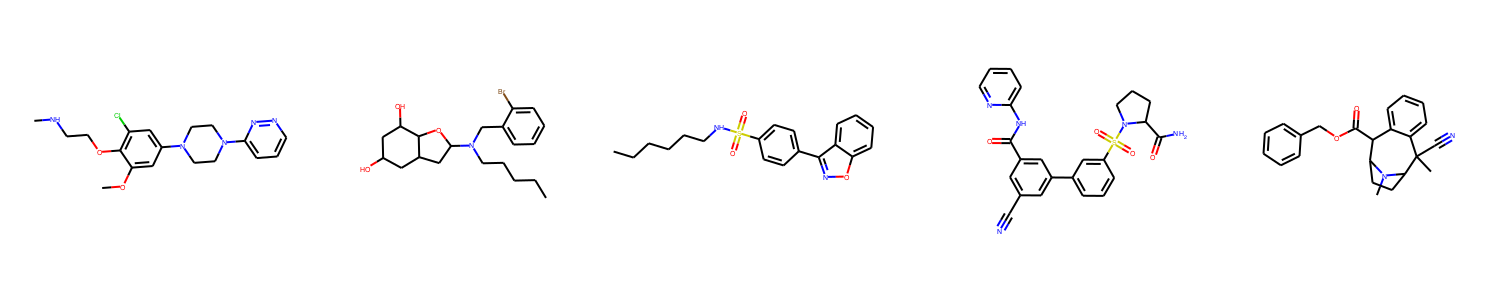}  
    \hfill
    \vspace{-15pt}
    \caption{Representative molecules generated for (top to bottom): NSP9 Replicase, Receptor-Binding Domain (RBD) of S protein, and  Main Protease of SARS-CoV-2}%
    \label{fig:generated_molecules}%
\end{figure*}

We show a representative set of molecules generated for each target in Figure ~\ref{fig:generated_molecules}

\FloatBarrier
\section{Novelty Analysis}
To assess the novelty of generated molecules, we assigned to each molecule $m$ a score $\mli{Nov}_m$ representing its minimal distance (maximal similarity) to all registered compounds $p$ in the training database or another reference database of known compounds $P$:$\mli{Nov}\_m = 1 - \max\_{p \in P}\{\textrm{sim}(k\_m, k\_p)\}$.
MACCS keys \cite{maccs_durant} were used as structural fingerprints to  determine the similarity
for each pair of molecule ($k_m$) and compound ($k_p$). 
The Tanimoto \cite{tanimoto_willett} coefficient between two fingerprints expresses the similarity:
$ \textrm{sim}(k_x, k_y) = \frac{|k_x \cap k_y|}{|k_x \cup k_y|}$.
Note that a novelty of 0 means that the molecule's fingerprint matches exactly the fingerprint of a compound in the reference database; however, the final structure of the generated molecule can still be different.

\begin{figure}[tbh]
    \centering 
    \includegraphics[width=0.9\textwidth]{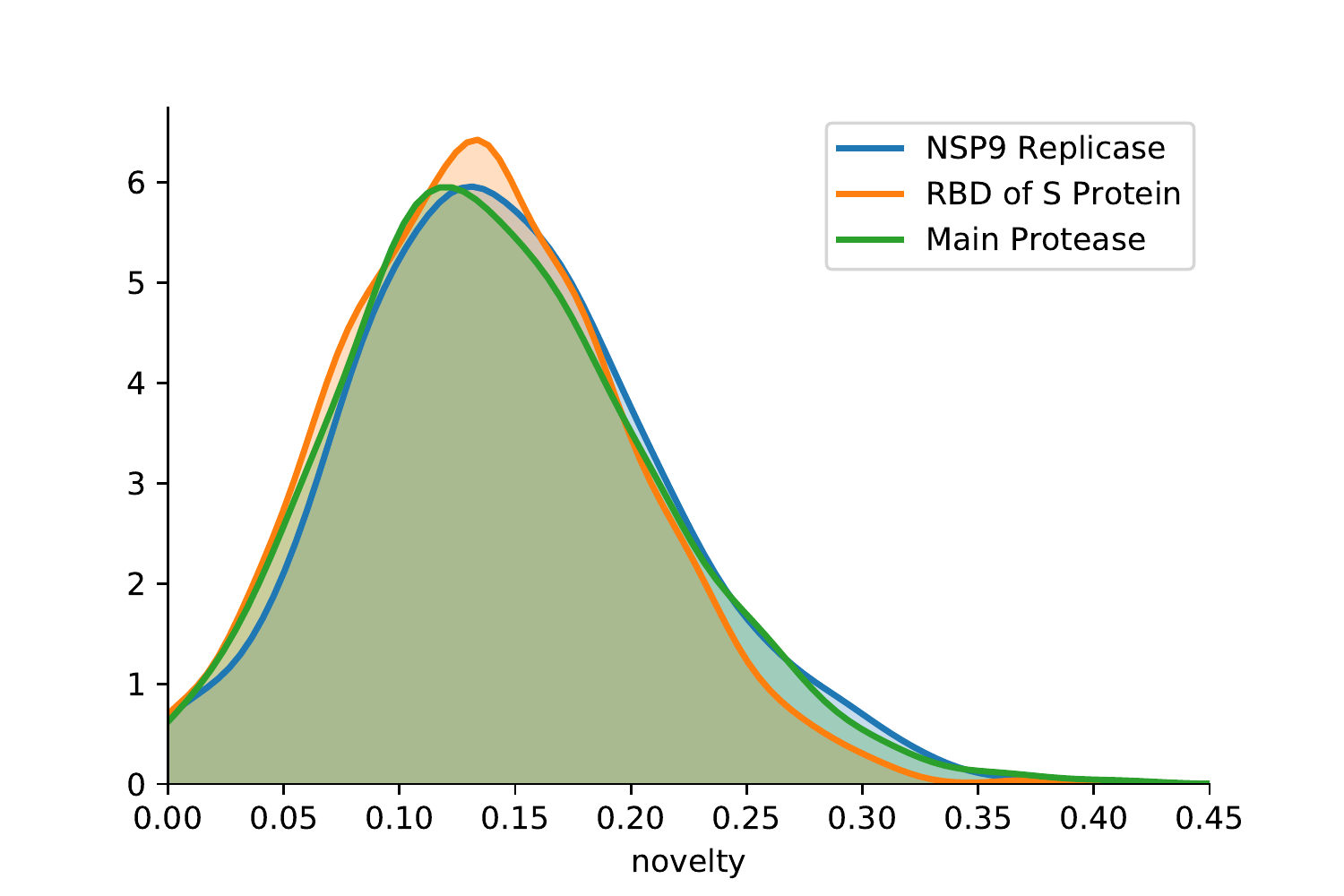}
    \caption{Novelty of the generated molecules for each target relative to the molecules in the training set, confirming that the model is indeed creating new molecules with novel fingerprints.}
    \label{fig:density_NOV_train}
  \end{figure}

\begin{figure}[tbh]
    \centering 
    \includegraphics[width=0.7\textwidth]{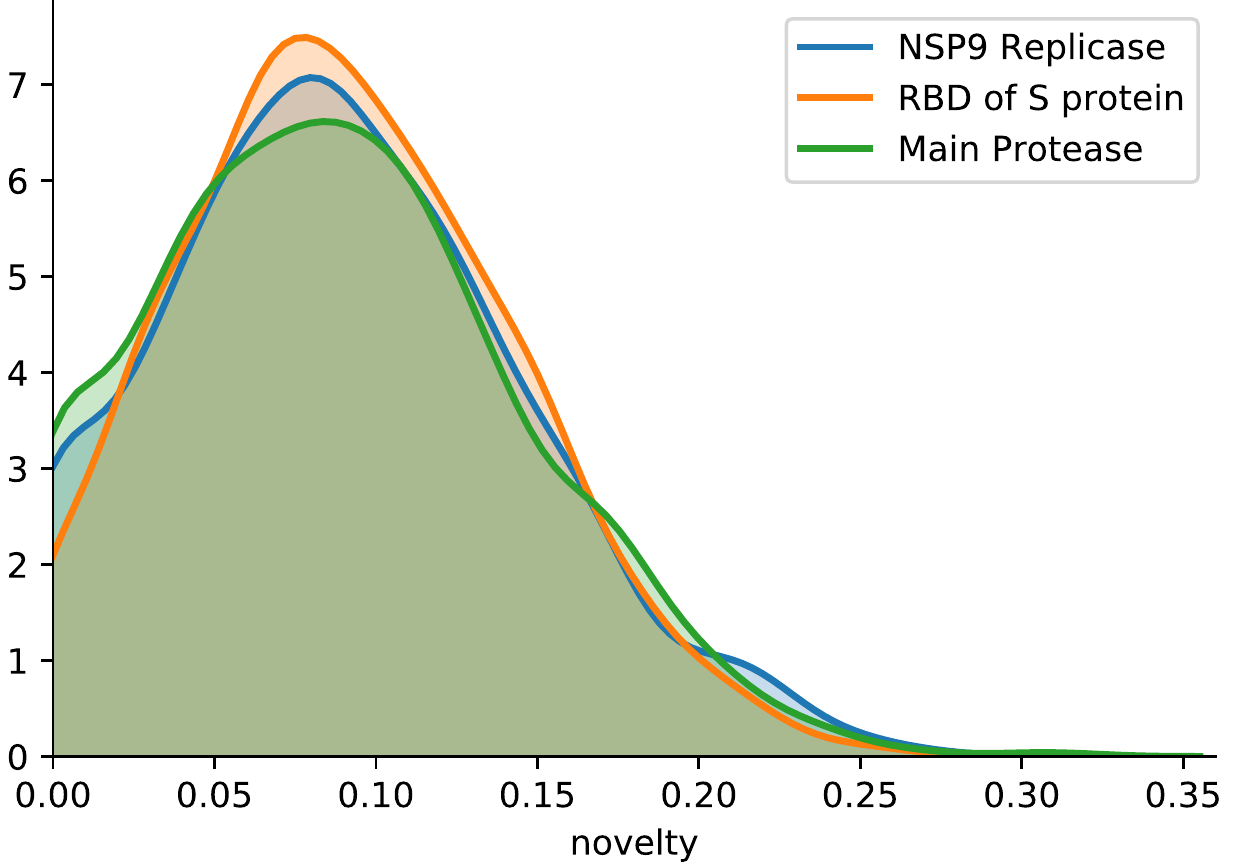}
\caption{Novelty relative to PubChem Dataset. Tends to be lower than novelty compared to all the training molecules, likely because there are significantly more molecules recorded in PubChem than the training subset.}
    \label{fig:density_NOV_pubchem}
  \end{figure}

The distribution of novelty scores for each of the targets with respect to the training set and a larger set of molecules from PubChem is given in Figures~\ref{fig:density_NOV_train} and ~\ref{fig:density_NOV_pubchem}.

We further compute the FCD of the generated molecules (and scaffolds) for each target with respect to the ZINC and BindingDB datasets (See Table \ref{SI:FCD}. We note that novel scaffolds emerge in the generated molecules with respect to both ZINC and BindingDB.



\begin{table}[ht]
\caption{FCD of the generated molecules for each target}
\scalebox{0.85}{
\begin{tabular}{
        r|        
        S[table-format=1.3]
        S[table-format=1.3]
        S[table-format=1.3]
        S[table-format=2.3]
        }
\toprule
{Target} & {FCD/Test (ZINC)} & {FCD/Test (BindingDB)} & {FCD/TestSF (ZINC)} & {FCD/TestSF(BindingDB)} \\
\midrule
NSP9   & 7.106 & 1.007 & 7.5 & 9.25  
\\ 
RBD    & 7.072 & 1.004 & 7.472 & 9.202 
\\ 
MPro   & 7.107 & 0.995 & 7.523 & 9.278 
\\ 
HDAC1  & 8.028 & 1.924 & 8.47 & 10.167  
\\ 
\bottomrule 
\end{tabular}
}
\label{SI:FCD}
\end{table}

\begin{figure}[htb]
  \centering
  \begin{subfigure}[htb]{0.475\textwidth}
    \centering
    \includegraphics[width=\textwidth]{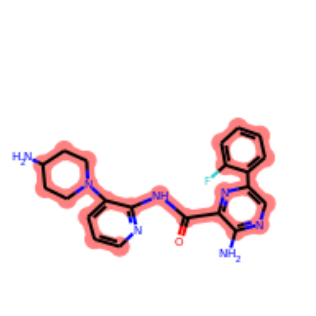}
  \end{subfigure}
  \hfill
  \begin{subfigure}[htb]{0.475\textwidth}  
    \centering 
    \includegraphics[width=\textwidth]{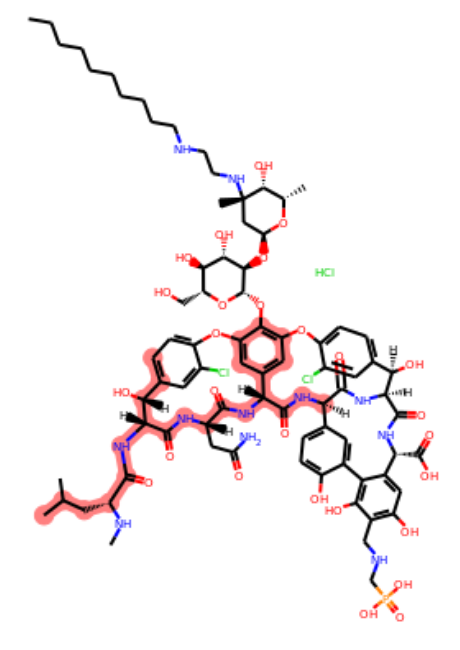}
    \label{fig:telavancin-sub}
  \end{subfigure}
  \caption{Maximum Common Subgraph Similarity of the CogMol-generated molecule with highest binding affinity to RBD (left) and Telavancin (right)}%
  \label{fig:telavancin}%
\end{figure}

It is also interesting to note that the CogMol generated molecule with the highest binding affinity to RBD has maximum subgraph similarity to a commercially available drug Telavancin (See Figure~\ref{fig:telavancin}). Telavancin is a  semi-synthetic derivative of vancomycin. It is used to treat complicated skin and skin structure infections, and  hospital-acquired and ventilator-associated bacterial pneumonia caused by Staphylococcus aureus.

\section{Property Predictors}
\label{SI:property_predictors}
Property predictors for QED, logP and SA were trained on the latent embeddings of the VAE. These regression models have 4 hidden layers with 50 units each and ReLU nonlinearity.
We further train a binding affinity predictor using the latent embeddings of the VAE and pretrained protein embeddings~\cite{Alley_2019}. 
The protein embeddings and the molecular embeddings are concatenated and passed through a single hidden layer with 2048 hidden units and ReLU nonlinearity. 
In order to build a predictor for binding affinity trained directly on the smiles sequences (x), we first embed them using LSTMs. When Protein sequences are embedded using LSTM, it serve as a baseline, and we get a RMSE of 1.0104 on test data. Our best model on Binding Affinity, with RMSE of 0.8426, uses pre-trained protein embeddings from~\cite{Alley_2019}.  

\begin{table}[ht]
    \centering
    \caption{
        Performance of the attribute predictors for QED, logP, SA, and binding affinities. Binding affinity (z) is trained on the latent space of the VAE, while binding affinity (x) is trained on the actual SMILES sequences.
    }
    \label{attr-predictor}
    \scalebox{0.85}{
        \begin{tabular}{rS[table-format=1.4]}
        \toprule
        {Attribute} & {RMSE} \\
        \midrule
        Binding Affinity (x) Baseline & 1.0104 \\ \hdashline
        QED & 0.0281 \\ 
        logP &  0.3307 \\ 
        SA & 0.0973 \\  
        Binding Affinity (z) &  1.2820 \\ 
        Binding Affinity (x) &  0.8426 \\ 
        \bottomrule
        \end{tabular}
    }
\end{table}

\section{Toxicity Prediction Model}
\label{SI:toxicity}

The MT-DNN model contains a total of four hidden layers: two are shared across all toxicity endpoints
and two are private for each of the endpoints.
We used a dropout \cite{Srivastava2014} probability of 0.5, and a 
ReLU activation function for all layers except for the last layers, in which the sigmoid activation was used. 
Morgan Fingerprints \cite{Rogers2010} were used as the input features to the model. 

The ROC AUC, Accuracy (ACC), Balanced Accuracy (BAC), True Negative (TN), True Positive (TP), Precision (PR), Recall (RC), and the F1 score of the MT-DNN model on Tox21 and ClinTox test data are reported in Table~\ref{tbl:tox_prediction} in the Appendix. Although the AUC values are slightly worse than the existing work of \cite[see Table S14]{Liu2019}, the precision (and thus true positive rate) achieved by the MT-DNN is much higher.\footnote{The average precision from \cite{Liu2019} over all 13 tasks is 0.45, which was obtained by running their code available through Github.} For comparison, we also report the results from a random forest (RF) model in Table~\ref{tbl:tox_RF}, showing that the MT-DNN significantly outperforms the RF model in terms of true positive rate, recall, and F1 score. Therefore, the MT-DNN model was used for assessing the generated molecules for toxicity.

Tables~\ref{tbl:tox_prediction} and~\ref{tbl:tox_RF} show the performance of toxicity prediction using the MT-DNN and the random forest as the baseline. The MT-DNN significantly outperforms the RF model in terms of true positive rate, recall, and F1 score, while incurring a small penalty in ROC AUC and precision. Table~\ref{tbl:tox_counts} displays the proportion of molecules being predicted toxic in a number of endpoints. We can see that the predicted toxicity of the generated molecules in all three targets are similar to that of the FDA approved drugs.


\input{tbl_tox_MT-DNN}
\input{tbl_tox_RF}
\input{tbl_tox_counts}
\FloatBarrier
\section{Docking Analysis}
\label{app:docking_analysis}

First, we removed chiral molecules from consideration for docking, as handling of chiral molecules in silico and in wet lab is tricky. We performed docking simulations using AutoDock Vina \cite{trott2010autodock}  with exhaustiveness=8 and a search space encompassing the entire protein target. We used the best result from 5 independent runs. Using a large set of approximately $875,000$ molecules (generated with only affinity constraints) for each target, we form clusters from the geometric centers of the top docking poses. We perform only 1 run of docking for these ligands. We use these cluster locations to approximate common binding sites for each target. We observe some correspondence with known binding pockets from literature --- e.g., cluster 0 for M\textsuperscript{pro} corresponds closely to the substrate-binding pocket \cite{zhang2020crystal} --- as well as with pockets identified with PrankWeb \cite{krivak2018p2rank,jendele2019prankweb} (see Table \ref{tbl:pocket_mapping}).

\begin{figure}[tbh]%
    \centering
    \begin{subfigure}[b]{\textwidth}
        \centering
        \includegraphics[height=0.25\textheight]{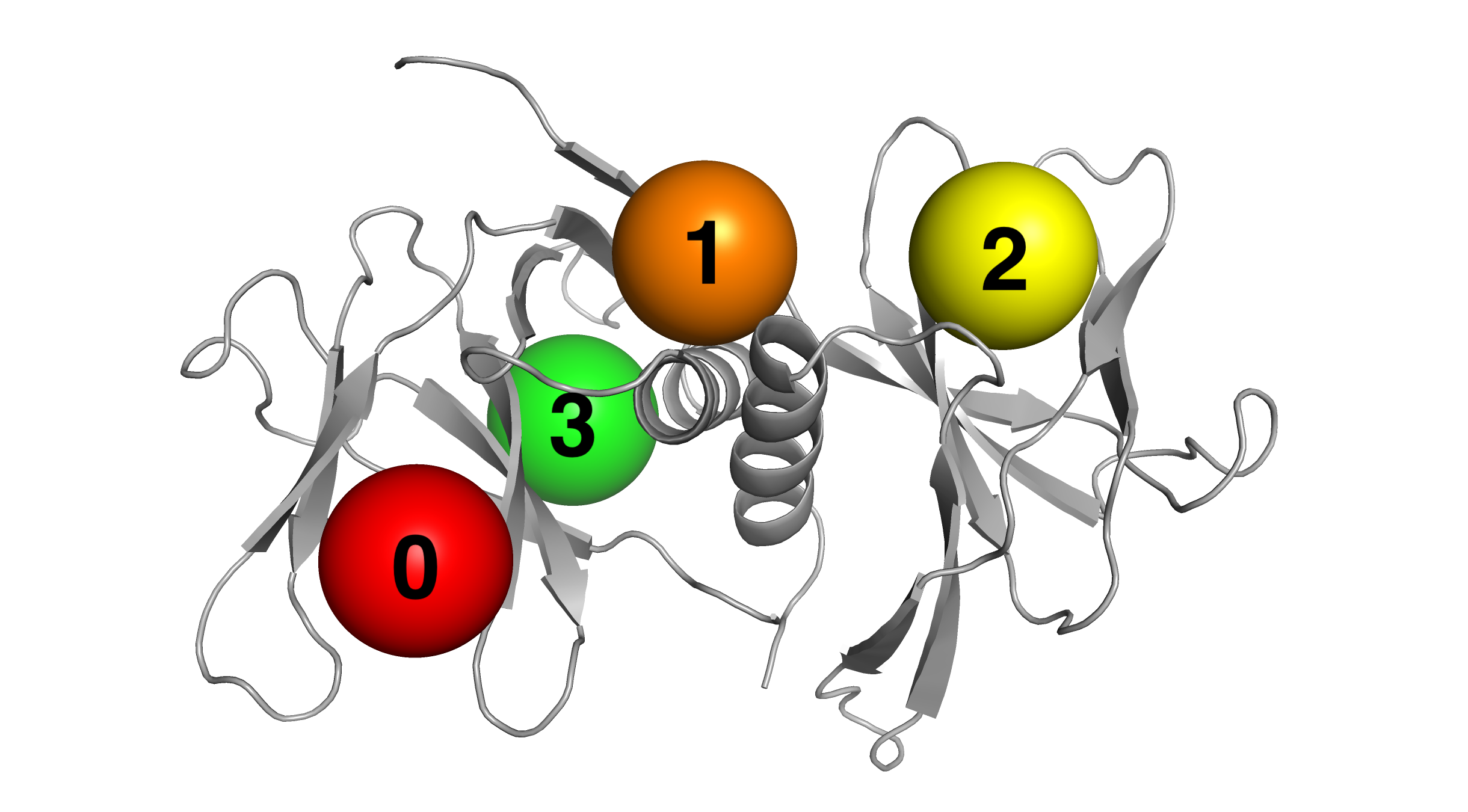}
        \caption{NSP9}
    \end{subfigure}
    \begin{subfigure}[b]{\textwidth}
        \centering
        \includegraphics[height=0.25\textheight]{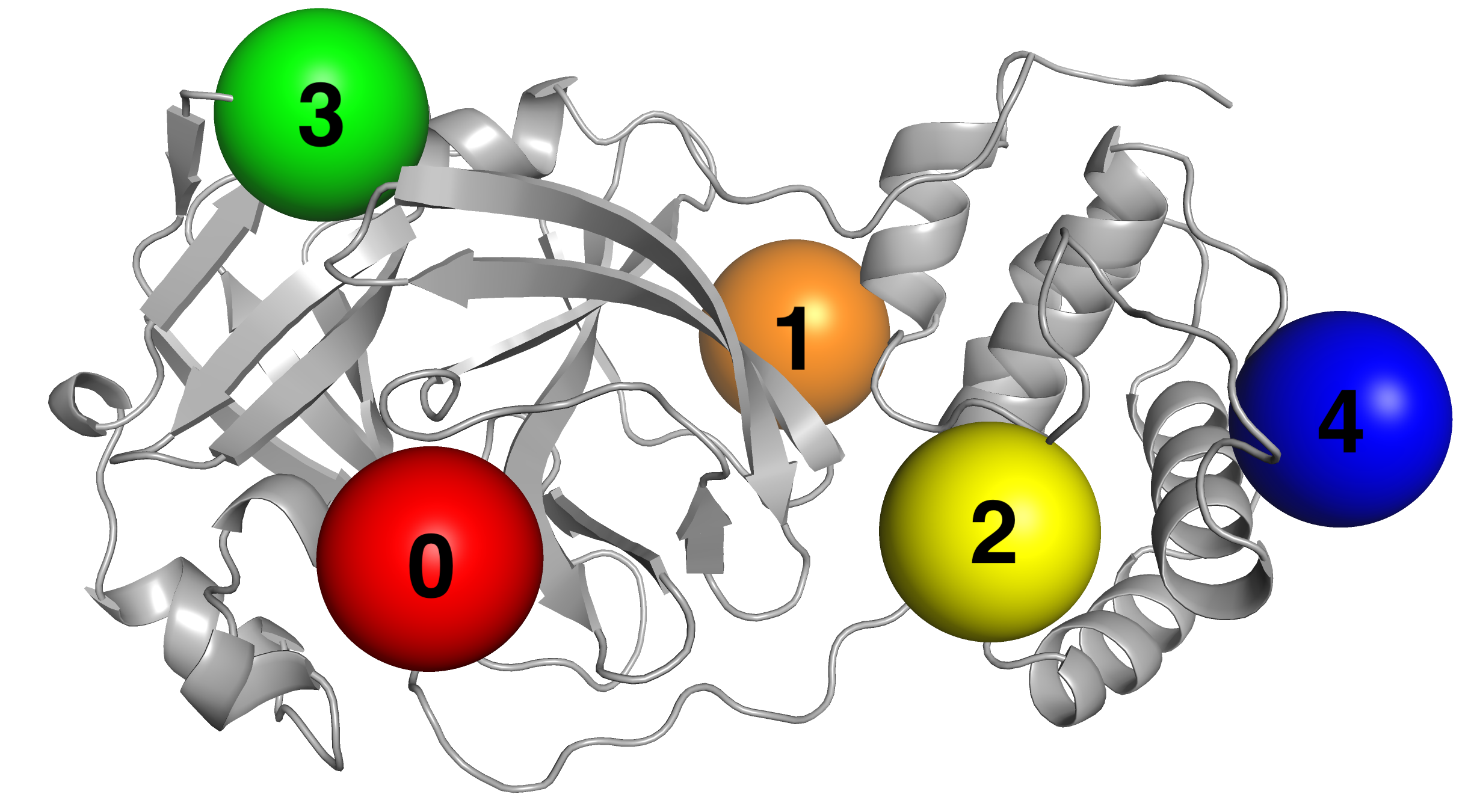}
        \caption{M\textsuperscript{pro}}
    \end{subfigure}
    \begin{subfigure}[b]{\textwidth}
        \centering
        \includegraphics[height=0.25\textheight]{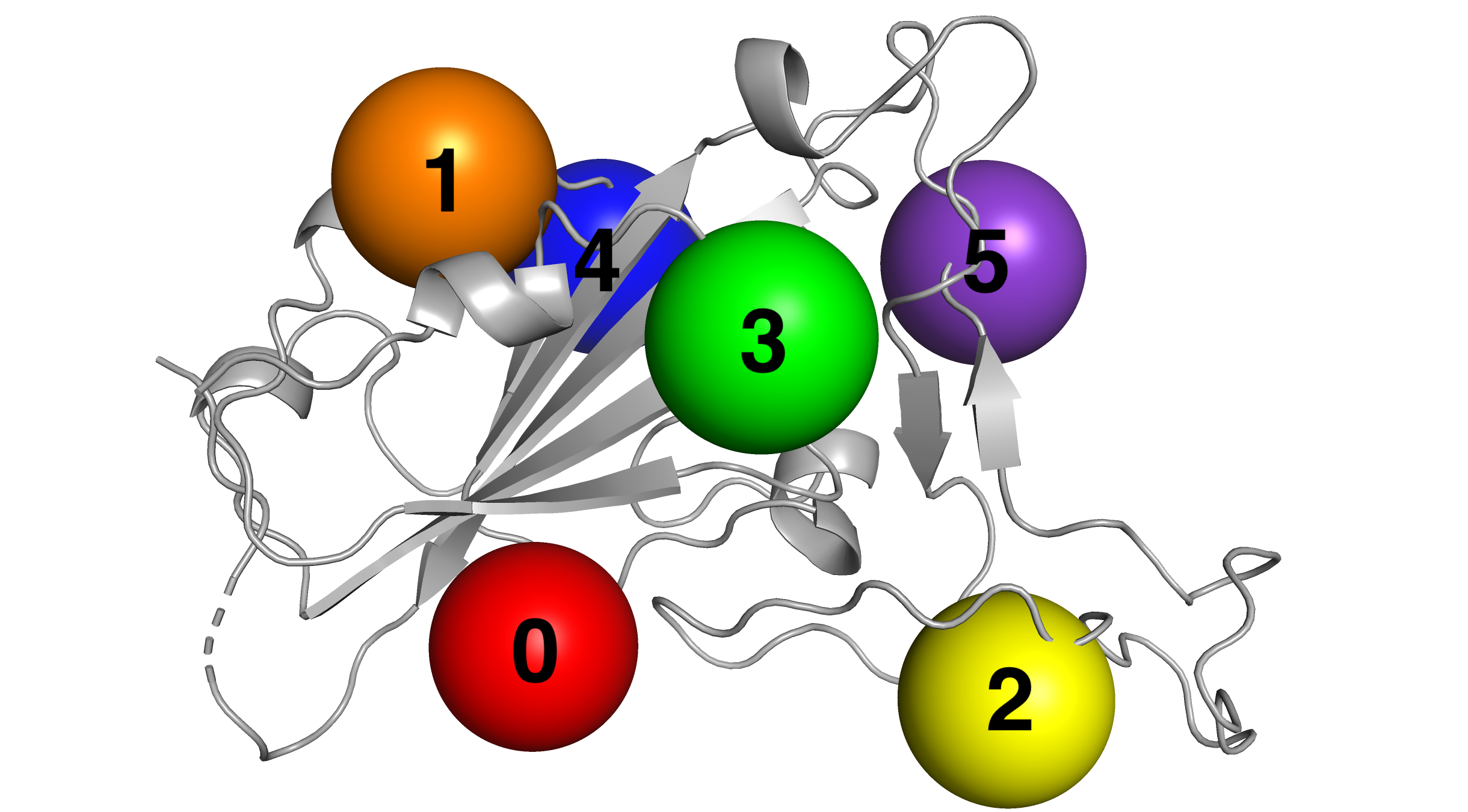}
        \caption{RBD}
    \end{subfigure}
    \vspace{10pt}
    \caption{Binding cluster mean locations from blind docking results for NSP9 Replicase, Main Protease, and receptor-Binding Domain (RBD) of  S protein of SARS-CoV-2.
    For each cluster, a large colored sphere is shown centered on the cluster mean. Note: these do not represent the exact extents of each cluster but simply serve to show the approximate locations.
    The indexes are used to refer to a specific cluster throughout this study. }%
    \label{fig:bindingclusters}%
\end{figure}

\begin{table}[tbh]
\centering
\caption{Docking analysis of \textit{screened} molecules: Size (\%), average ($\pm$ standard deviation) binding free energy, minimum binding free energy, the fraction of generated molecules with binding free energy $<-6$ kcal/mol for each cluster. In parentheses after target name: \% of  generated molecules  that showed a binding free energy $<-6$ kcal/mol. Table \ref{tbl:docking_1K} shows a condensed version of this table.}
\label{tbl:docking_1K_full}
\scalebox{0.85}{
\begin{tabular}{rl|
                S[table-number-alignment=center]
                S[table-number-alignment=center,separate-uncertainty,table-figures-uncertainty=1,table-figures-decimal=1]
                S[table-number-alignment=center,table-format=1.1]
                S[table-number-alignment=center]}
\toprule
\multicolumn{2}{c|}{Target} & {Size (\%)} & {Mean (kcal/mol)} & {Min (kcal/mol)} & {Low Energy (\%)} \\
\midrule
\multirow{4}{*}{NSP9 Dimer (87\%)} & cluster 0 & 67 & -6.8 \pm 0.7 & -8.6 & 88\\
& cluster 1 & 22 & -6.9 \pm 0.9 & -8.8 & 85\\
& cluster 2 & 9  & -7.0 \pm 0.8 & -8.8 & 80\\
& cluster 3 & 3  & -6.5 \pm 0.9 & -8.1 & 73\\[10pt]

\multirow{4}{*}{Main Protease (91\%)} & cluster 0 & 76 & -7.2 \pm 0.8 & -9.5 & 93\\
& cluster 1 & 18 & -6.9 \pm 0.8 & -9.2 & 86\\
& cluster 2 & 4  & -7.0 \pm 0.5 & -7.8 & 94\\
& cluster 3 & 2  & -7.0 \pm 1.0 & -8.4 & 75\\
& cluster 4 & 1  & -6.8 \pm 1.3 & -8.2 & 67\\[10pt]

\multirow{6}{*}{RBD (95\%)} & cluster 0 & 30 & -6.9 \pm 0.6 & -8.3 & 93\\
& cluster 1 & 36 & -7.2 \pm 0.6 & -9.1 & 97\\
& cluster 2 & 18 & -6.8 \pm 0.7 & -8.3 & 84\\
& cluster 3 & 12 & -7.3 \pm 0.5 & -9.1 & 100\\
& cluster 4 & 3  & -6.9 \pm 0.7 & -8.0 & 86\\
& cluster 5 & 1  & -6.8 \pm 0.4 & -7.3 & 100\\
\bottomrule
\end{tabular}}
\end{table}

\begin{table}[b]
\centering
\caption{Summary of docking for larger set of generated molecules: cluster size as a percentage of total molecules per target, average ($\pm$ standard deviation) binding free energy, minimum binding free energy, the fraction of generated molecules with binding free energy $\leq -6$ kcal/mol. In parentheses after target name: 86\%, 90\%, and 84\% of all generated molecules for the respective targets showed a binding free energy of $\leq -6$ kcal/mol. The molecules used were generated with just an affinity criterion and \textit{not screened} for toxicity or retrosynthesis. These were used to fit the Gaussian Mixture Models and form clusters shown in Figure \ref{fig:bindingclusters}.}
\label{tbl:docking_1M}
\scalebox{0.85}{
\begin{tabular}{rl|
                S[table-number-alignment=center]
                S[table-number-alignment=center,separate-uncertainty,table-figures-uncertainty=1,table-figures-decimal=1]
                S[table-number-alignment=center,table-format=1.1]
                S[table-number-alignment=center]}
\toprule
\multicolumn{2}{c|}{Target} & {Size (\%)} & {Mean (kcal/mol)} & {Min (kcal/mol)} & {Low Energy (\%)} \\
\midrule
\multirow{4}{*}{NSP9 Dimer (86\%)} & cluster 0 & 70 & -6.9 \pm 0.8 & -10.7 & 87\\
& cluster 1 & 18 & -6.9 \pm 0.9 & -12.0 & 85\\
& cluster 2 & 9 & -7.0 \pm 0.8 & -10.5 & 86\\
& cluster 3 & 4 & -6.6 \pm 0.8 & -9.8 & 75\\[10pt]

\multirow{4}{*}{Main Protease (90\%)} & cluster 0 & 58 & -7.2 \pm 0.8 & -10.7 & 92\\
& cluster 1 & 26 & -7.1 \pm 0.9 & -10.9 & 88\\
& cluster 2 & 11 & -7.1 \pm 0.8 & -10.9 & 91\\
& cluster 3 & 2 & -7.1 \pm 0.8 & -9.9 & 89\\
& cluster 4 &2 & -6.4 \pm 1.0 & -10.0 & 63\\[10pt]

\multirow{6}{*}{RBD (84\%)} & cluster 0 & 40 & -6.7 \pm 0.8 & -10.6 & 81\\
& cluster 1 & 28 & -7.0 \pm 0.8 & -12.4 & 90\\
& cluster 2 & 16 & -6.6 \pm 0.7 & -9.6 & 77\\
& cluster 3 &11 & -7.1 \pm 0.7 & -10.7 & 93\\
& cluster 4 &3 & -6.7 \pm 0.8 & -9.8 & 78\\
& cluster 5 &1 & -6.9 \pm 0.7 & -10.5 & 89\\
\bottomrule
\end{tabular}}
\end{table}

\begin{table}[ht]
\centering
\caption{Cluster mappings for identified binding pockets. Pockets are ordered according to descending score (a combination of predicted ligandability and conservation). Note: clusters encompass larger regions so multiple pockets may correspond to the same cluster.}
\label{tbl:pocket_mapping}
\scalebox{0.85}{
\begin{tabular}{rl|r}
\toprule
\multicolumn{2}{c|}{Target} & Cluster \\
\midrule
\multirow{4}{*}{NSP9 Dimer} & pocket 0 & 2\\
& pocket 1 & 0\\
& pocket 2 & 1\\
& pocket 3 & 1\\[10pt]

\multirow{4}{*}{Main Protease (91\%)} & pocket 0 & 0\\
& pocket 1 & 1\\
& pocket 2 & 1\\
& pocket 3 & 3\\[10pt]

\multirow{6}{*}{RBD (95\%)} & pocket 0 & 5\\
& pocket 1 & 4\\
& pocket 2 & 0\\
& pocket 3 & 1\\
& pocket 4 & 2\\
& pocket 5 & 3\\
\bottomrule
\end{tabular}}
\end{table}

\begin{figure}[bth]
    \begin{subfigure}[b]{\textwidth}
        \centering
        \includegraphics[height=0.25\textheight]{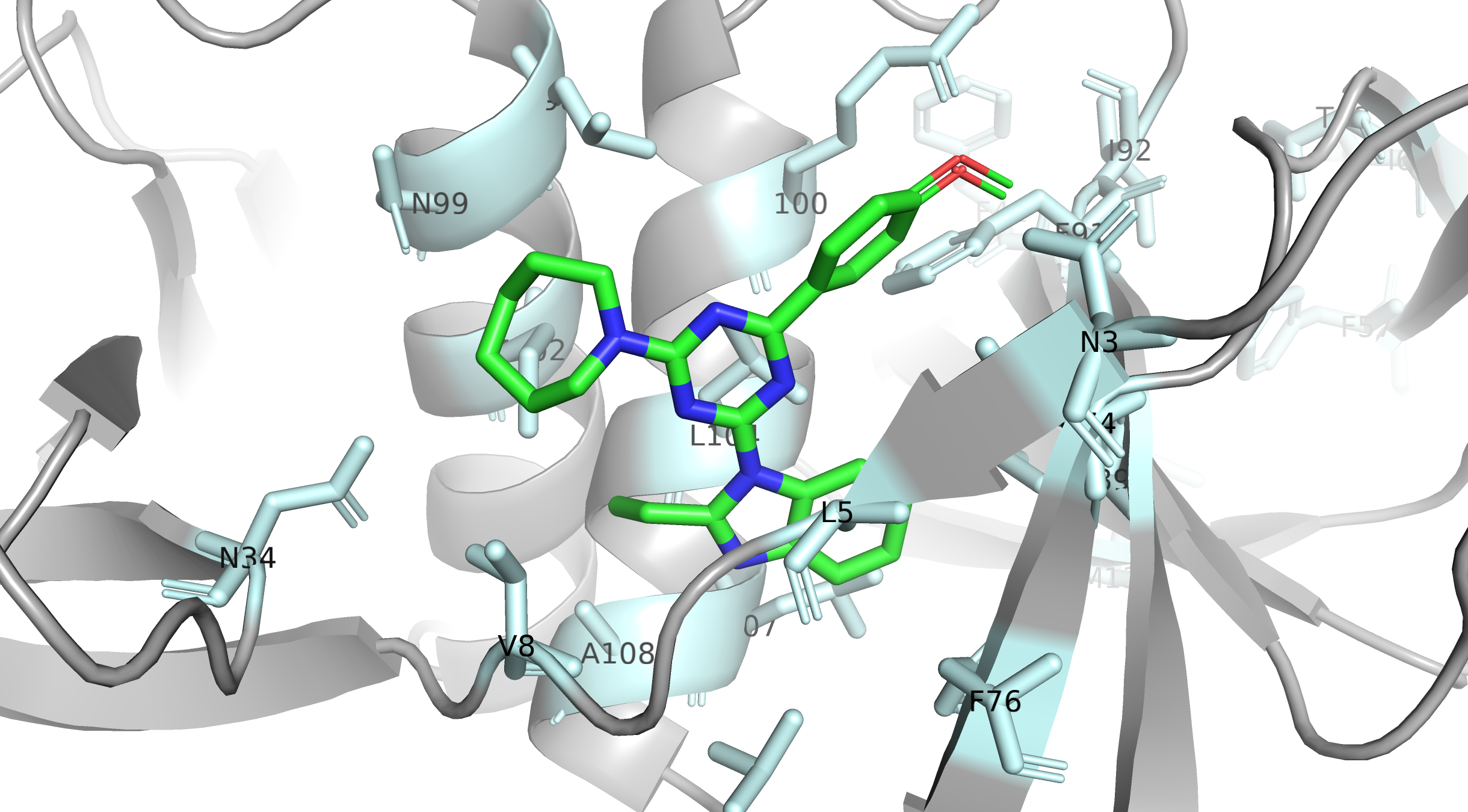}
        \caption{NSP9 Dimer (cluster 1)}
        \label{fig:nsp9_binding}
    \end{subfigure}
    \vskip\baselineskip
    \begin{subfigure}[b]{\textwidth}
        \centering
        \includegraphics[height=0.25\textheight]{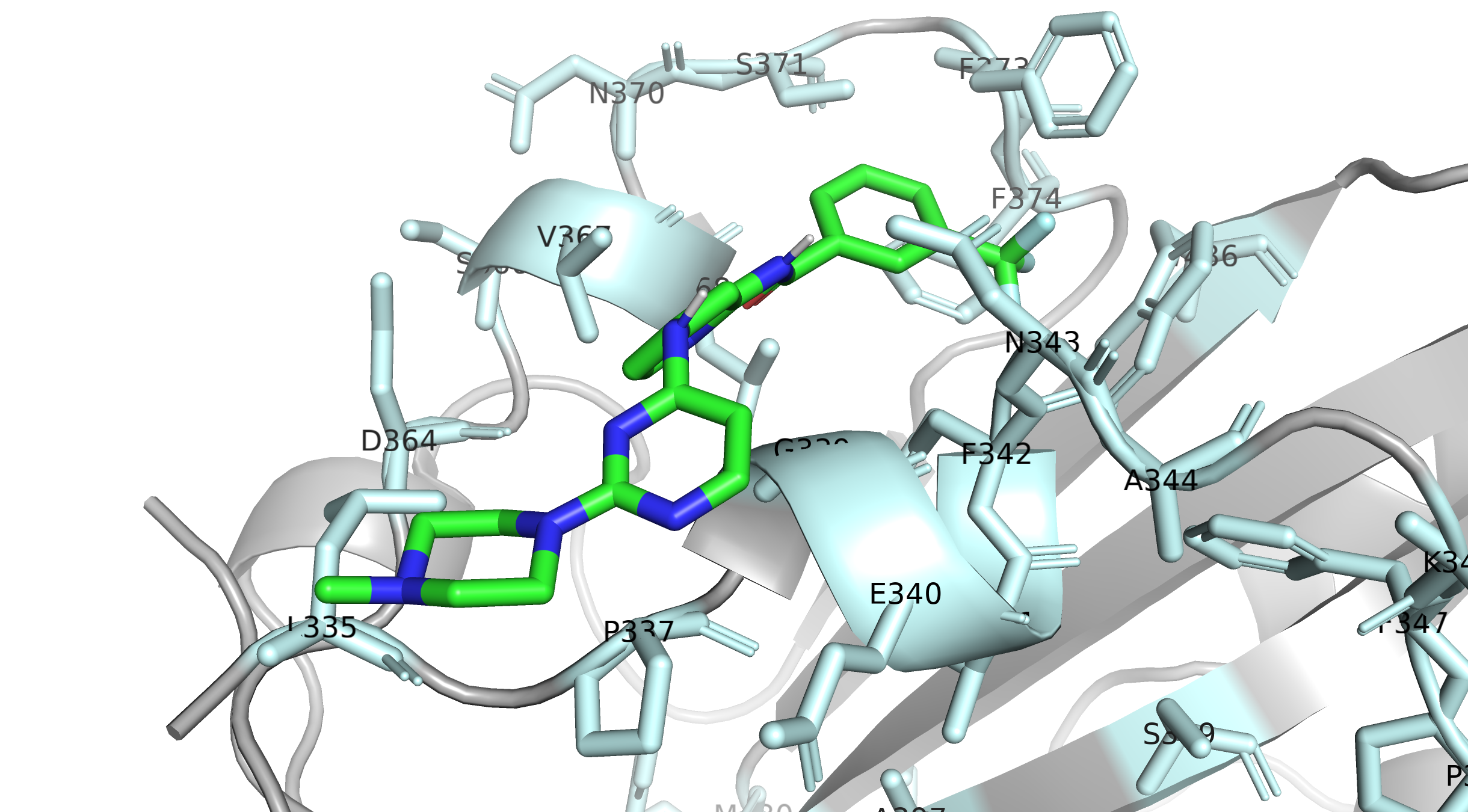}
        \caption{RBD (cluster 1)}
    \label{fig:rbd_binding}
    \end{subfigure}
    \vskip\baselineskip
    \begin{subfigure}[b]{\textwidth}
        \centering
        \includegraphics[height=0.25\textheight]{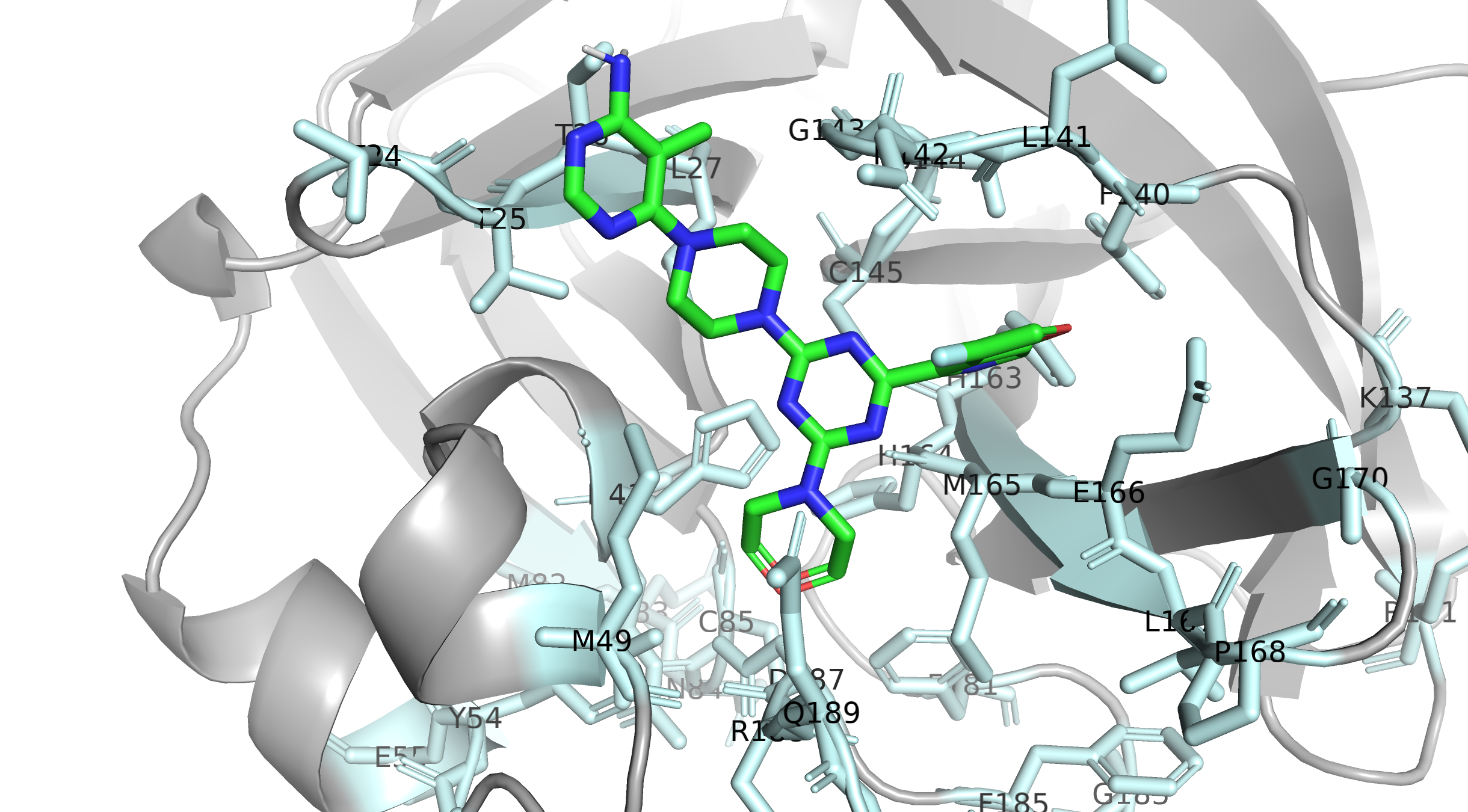}
        \caption{Main Protease (cluster 0)}
        \label{fig:main_protease_binding}
    \end{subfigure}
    \caption{Predicted lowest energy binding mode for top generated molecule (in terms of docking binding free energy) with the corresponding target. Protein residues within 4 angstroms of the generated ligand are shown in cyan.}
    \label{fig:docking}
\end{figure}

\begin{figure}[tbh]
    \centering
    \setlength{\fboxsep}{0pt}
    \includegraphics[width=\textwidth]{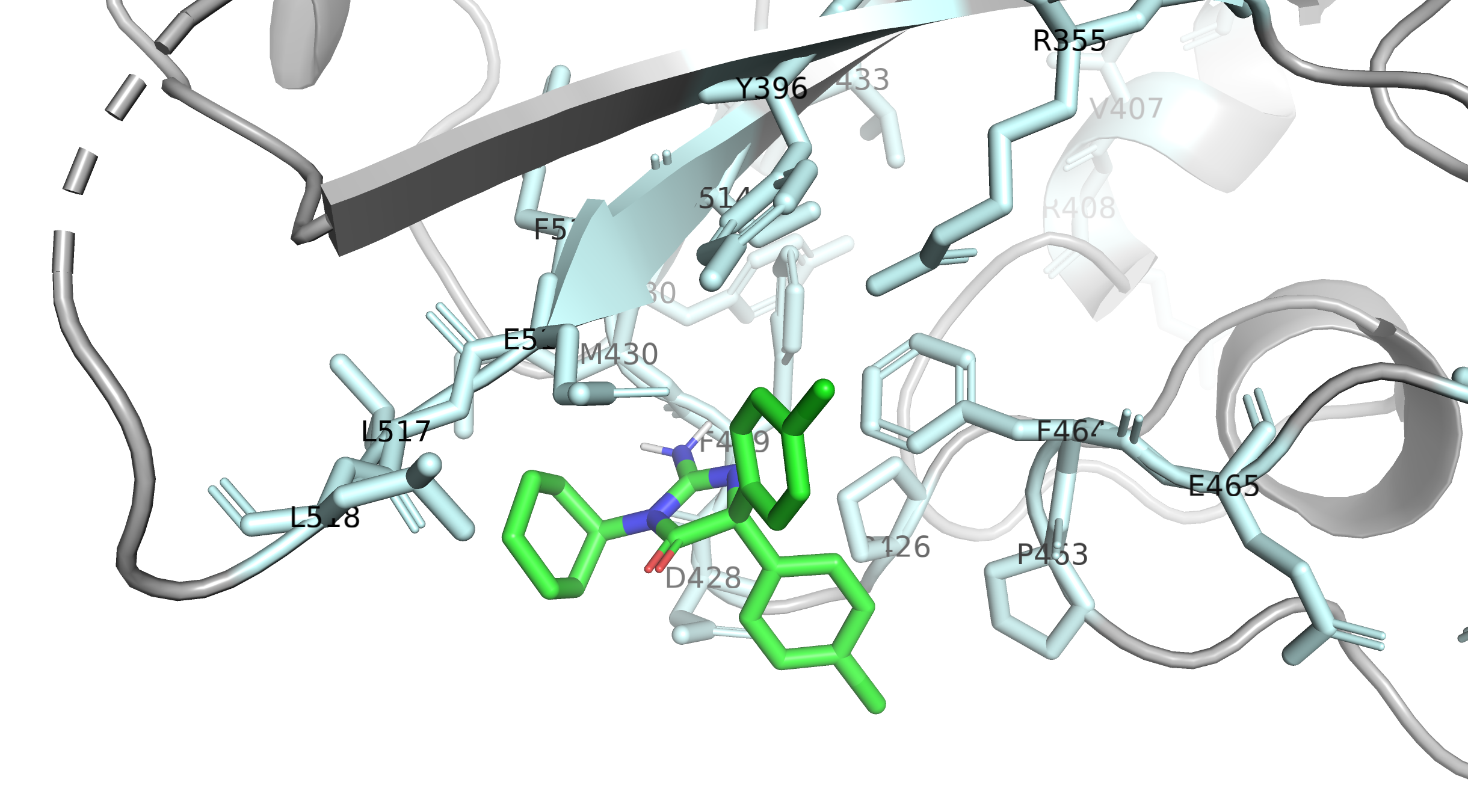}
    \caption{Predicted lowest energy binding mode for PubChem Compound ID 76332092 with RBD (cluster 0).}
    \label{fig:rbd_GEN146_docking}
\end{figure}

\FloatBarrier

\section{Additional Details and Analysis of the Retrosynthesis}
\label{SI:retrosynthesis}

The generated molecules were tested for synthesizability, using a retrosynthetic algorithm~\cite{schwaller2020predicting}\footnote{The predictions have been performed using the python package \texttt{rxn4chemistry} (\url{https://github.com/rxn4chemistry/rxn4chemistry})} based on the Molecular Transformer~\cite{SchwallerFWD}, trained on patent chemical reaction data.
The performance of the retrosynthetic model depends on the type of chemistry needed to design a specific compound.
Indeed, while chemical reactions from patents contain a wide variety of synthetic strategies, their variability is strongly biased by those reaction schemes that are mostly used in pharma and chemical industry.
The need of academic type of reactions in the synthesis of a molecule may result in the entire design being not successful due to the lack of specific chemical knowledge.
It is also important to remark that the retrosynthetic model~\cite{schwaller2020predicting} does not memorize data, but captures and learns chemical reaction patterns.
The retrosynthesis is considered successful when, within the maximum allowed number of steps, the algorithm is able to reach commercially available materials 
All retrosynthesis have been executed with a maximum number of retrosynthetic steps equal to 6, a forward acceptance probability of 0.6, number of beams equal to 10, pruning steps equal to 2, no precursor price threshold and maximum execution time of 1 hour\footnote{For a detailed description of all the parameters see~\cite{schwaller2020predicting} or \texttt{rxn4chemistry} docs:
\url{https://rxn4chemistry.github.io/rxn4chemistry/\_modules/rxn4chemistry.html\#rxn4chemistry.core.RXN4ChemistryWrapper.predict\_automatic\_retrosynthesis}}. All commercially available products have been extracted from the emolecules database~\cite{eMolecules} restricting the selection to materials with a lead time of 4 weeks or less.

An important note on the FDA set considered in the analysis~\cite{FDAenamine}, it that among the 682 compounds only 489 molecules have been considered in the analysis after removal of entries mapping multiple chemicals, such as salts.

\begin{figure}[!htb]
    \centering
    \includegraphics[width=0.75\textwidth]{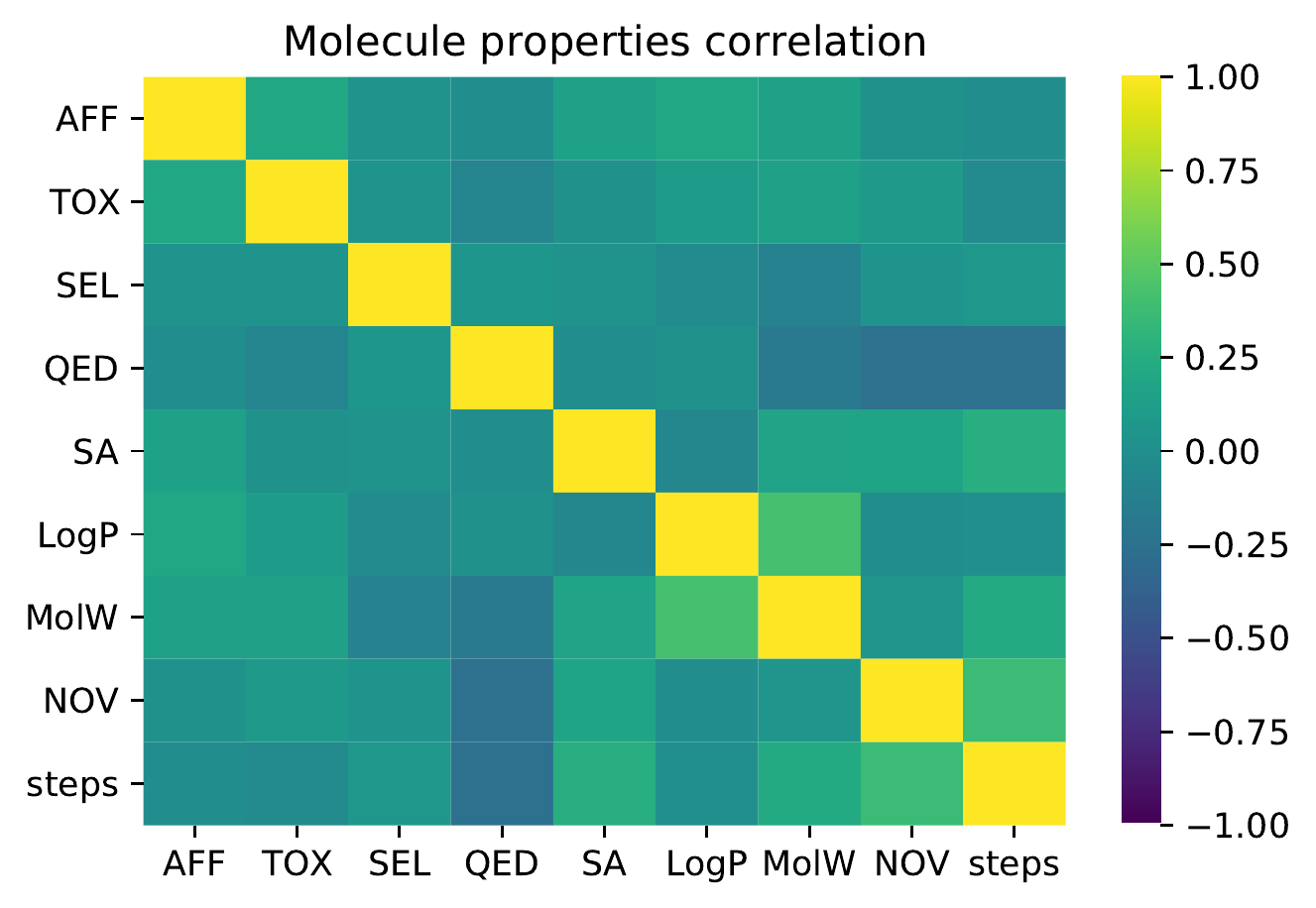}
    \caption{Main Protease correlation between properties and synthesis.}
    \label{fig:correlation_MPRO}
\end{figure}

\begin{figure}[!htb]
    \centering
    \includegraphics[width=0.75\textwidth]{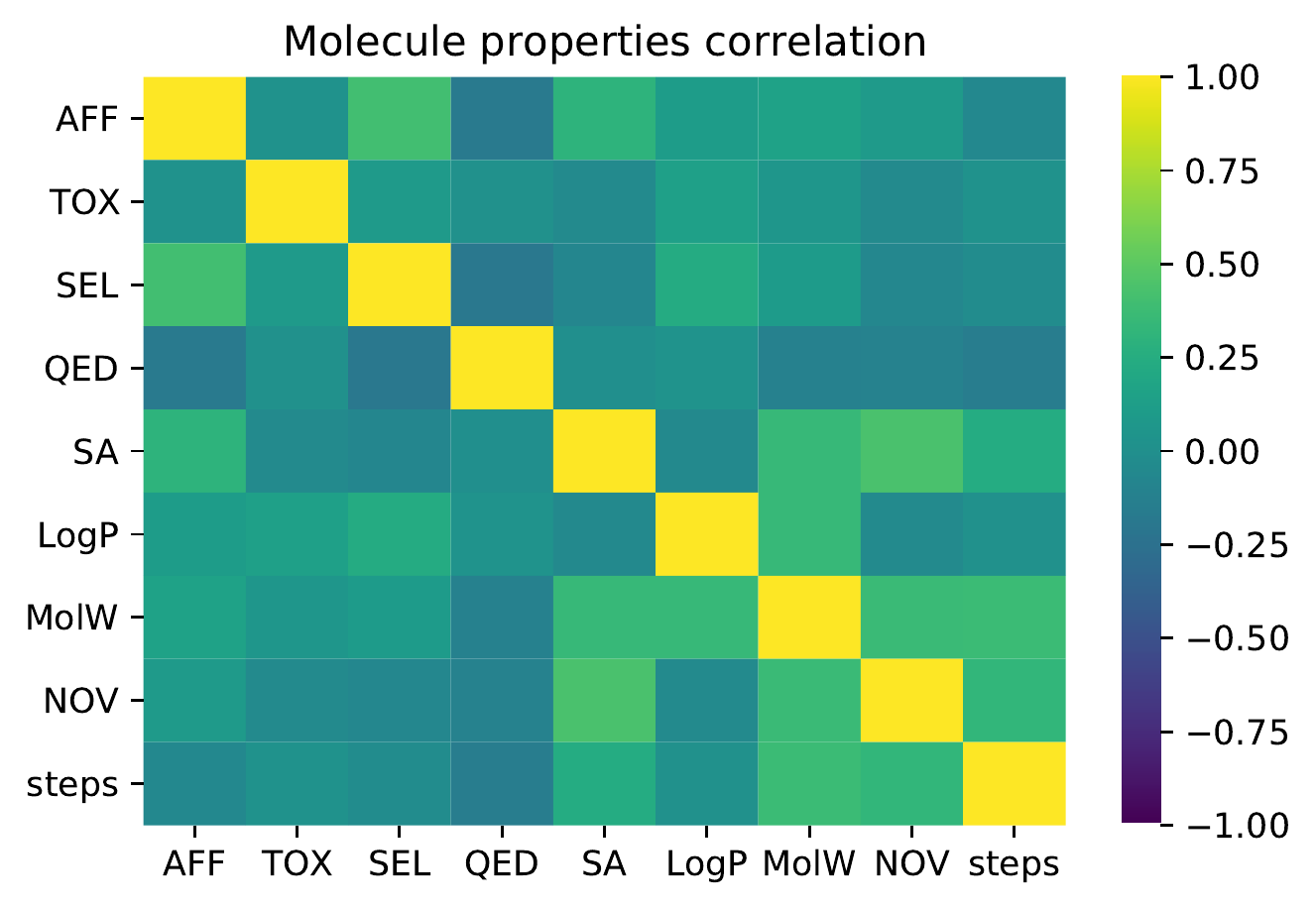}
    \caption{NSP9 Replicase correlation between molecule properties and synthesis.}
    \label{fig:correlation_NSP9}
\end{figure}

\begin{figure}[!htb]
    \centering
    \includegraphics[width=0.75\textwidth]{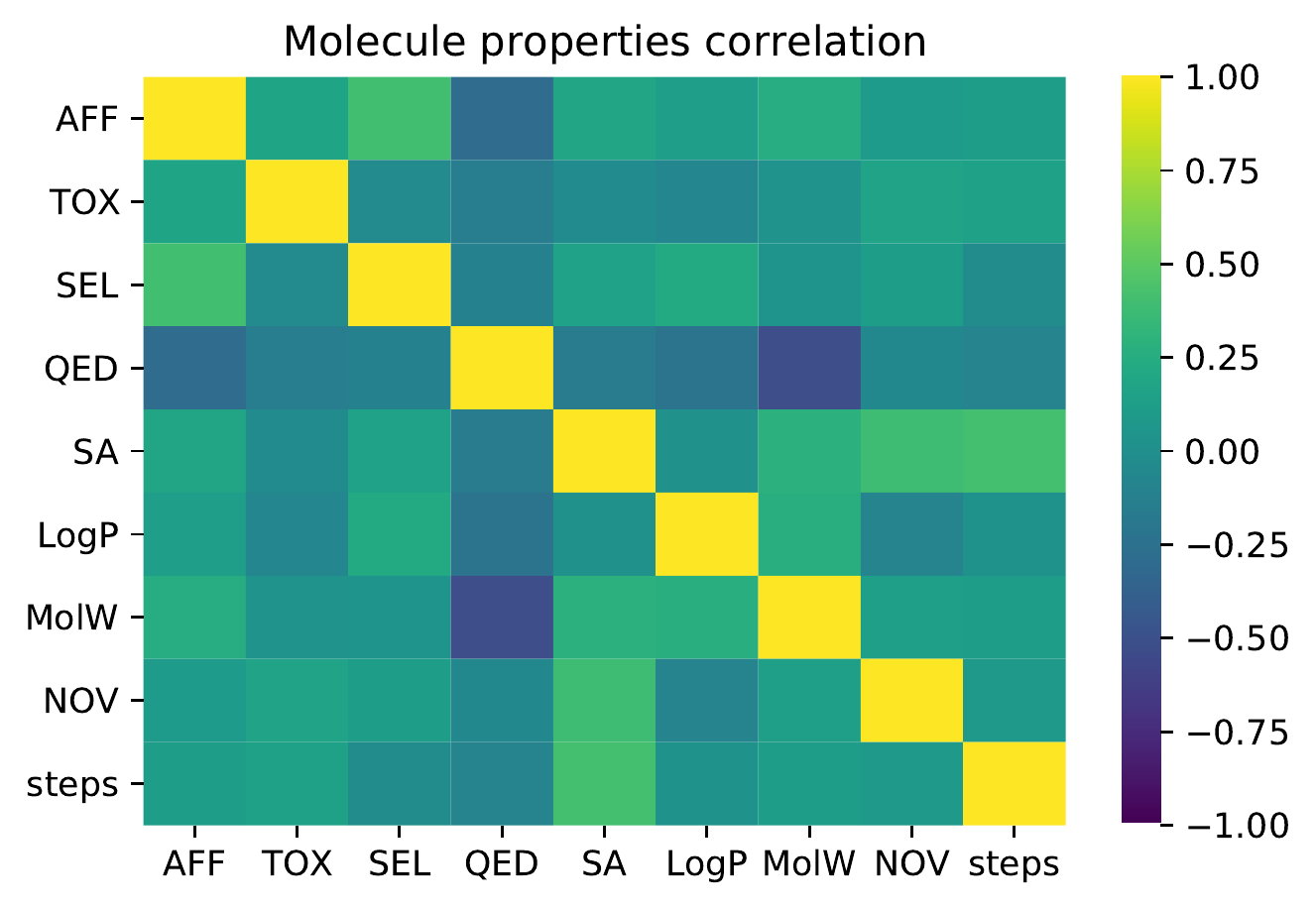}
    \caption{RBD of S Protein correlation between molecule properties and synthesis.}
    \label{fig:correlation_RBD}
\end{figure}

The reported number of steps (steps) was used to build correlation plots (see Figures~\ref{fig:correlation_MPRO},\ref{fig:correlation_NSP9},\ref{fig:correlation_RBD}) among a large number of descriptors, including: a measure of the affinity (AFF), toxicity (TOX), selectivity (SEL), drug-likeness (QED), synthetic accessibility (SA), partition coefficient (LogP), the molecular weight (MolW), novelty (NOV). We report the correlation analysis for the M\textsuperscript{pro} (see Figures~\ref{fig:correlation_MPRO}), NSP9 (see Figures~\ref{fig:correlation_NSP9}) and RBD target (see Figures~\ref{fig:correlation_RBD}).
As expected, we observe a positive correlation between SA and the number of steps.
It is interesting to observe how the novelty (NOV) is positively correlated with the number of steps for both the Main Protease and the NSP9 Replicase, while for the RBD there is no evident correlation.
This confirms that in the latter case, while the scaffolds are as novel as the M\textsuperscript{pro} and the NSP9-related molecules, the type of precursors needed exhibit a greater variability of chemical complexity resulting in an uncorrelated number of steps needed to complete the synthesis.

\FloatBarrier
\section{Molecule Explorer}
\label{SI:molexplorer}
\begin{figure}[!ht] 
    \centering
    \includegraphics[width=0.9\textwidth]{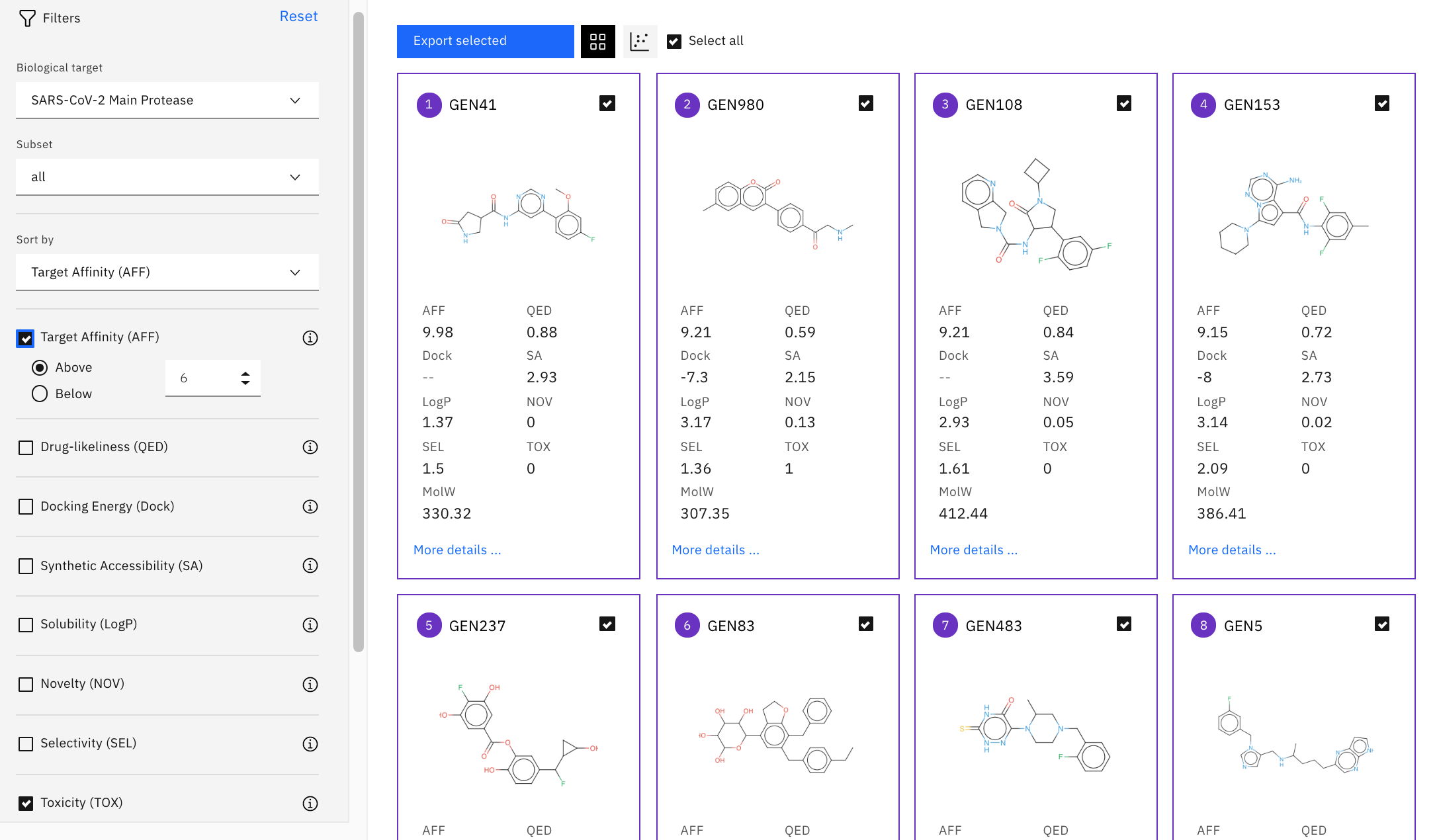}
    \caption{Generated molecules for Main Protease with high affinity and low toxicity are displayed in a list.}
    \label{fig:demo_main_protease_list}
\end{figure}

\begin{figure}[!ht] 
    \centering
    \includegraphics[width=0.9\textwidth]{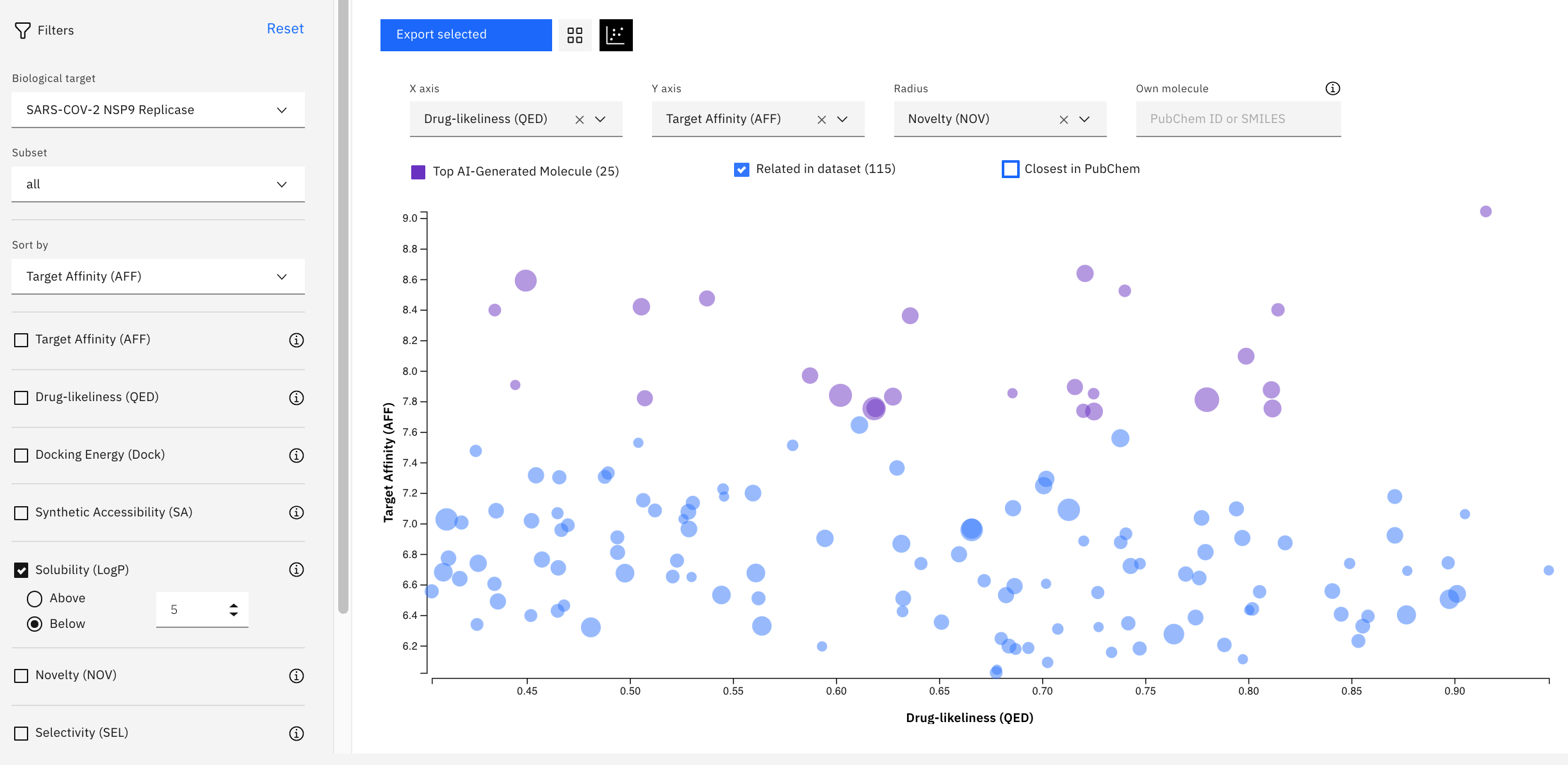}
    \caption{Generated molecules for NSP9 with low solubility and low toxicity, displayed in a plot of target affinity vs drug likeliness (QED).}
    \label{fig:demo_nsp9_plot}
\end{figure}

\begin{figure}[!ht] 
    \centering
    \includegraphics[width=0.9\textwidth]{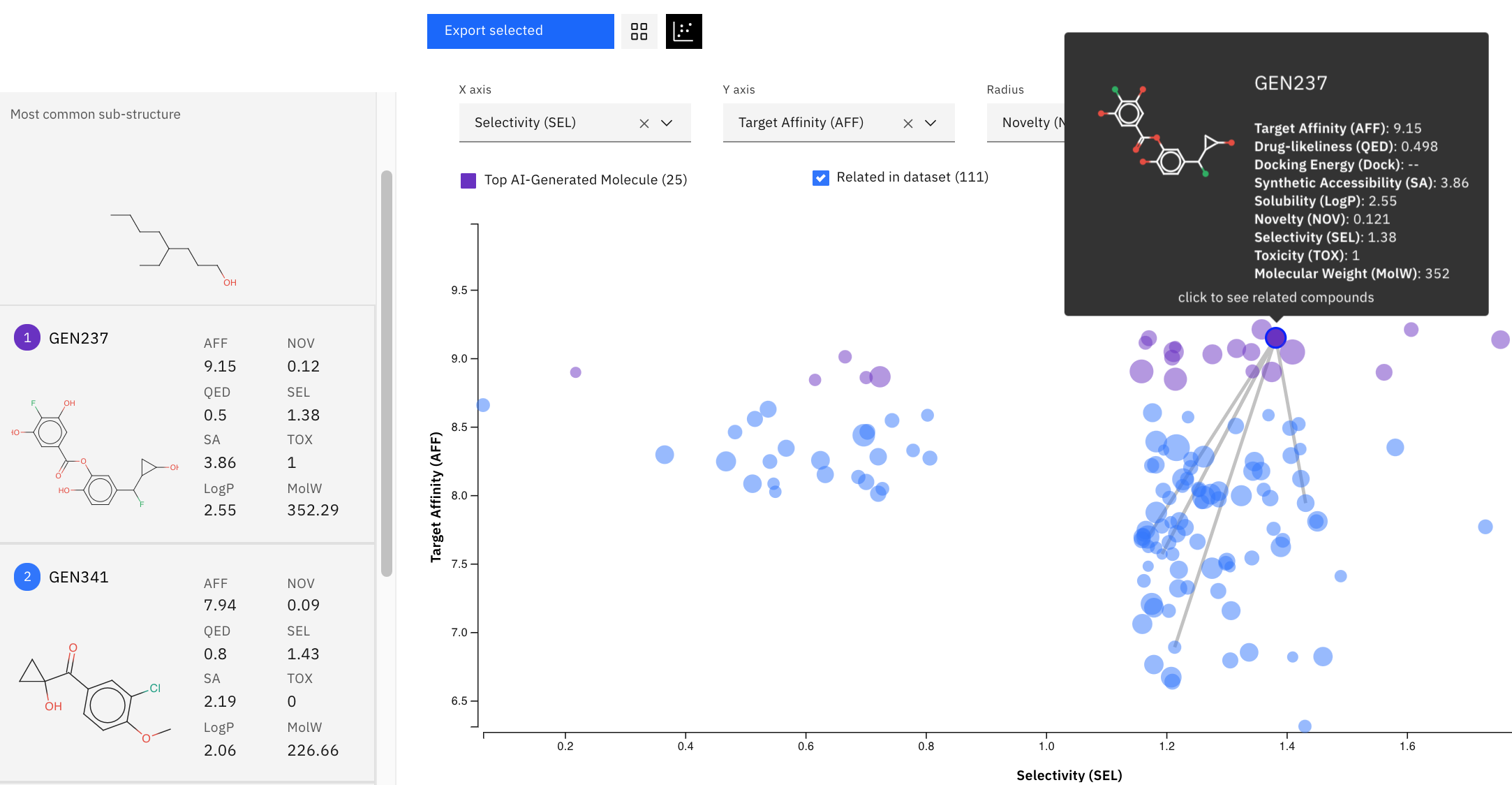}
    \caption{Generated molecules for Main Protease displayed in a plot format, here, the related molecules for GEN237 are shown via lines to those in the dataset. Sub-structure common to the molecules are highlighted on the left, with examples listed.}
    \label{fig:demo_main_protease_related}
\end{figure}

In order to allow domain experts to explore and screen the generated molecules for further analysis, we created an intuitive Molecule Explorer tool. The tools allows a user to filter molecules based on a variety of properties (QED, Target Affinity, Docking Energy, Synthetic Accessibility, Selectivity, Solubility and Novelty) and identify existing molecules in PubChem that are closest to the generated molecule.
We present some screenshots of the molecule explorer in Figures~\ref{fig:demo_main_protease_list}, \ref{fig:demo_nsp9_plot}, and \ref{fig:demo_main_protease_related}. A screencast of the tool can be viewed at \url{https://www.youtube.com/watch?v=cYb8_catBpI}

\section{HDAC1: Molecule Generation for a Target with Low Coverage}
\begin{figure}[!ht] 
    \centering
    \includegraphics[width=0.5\textwidth]{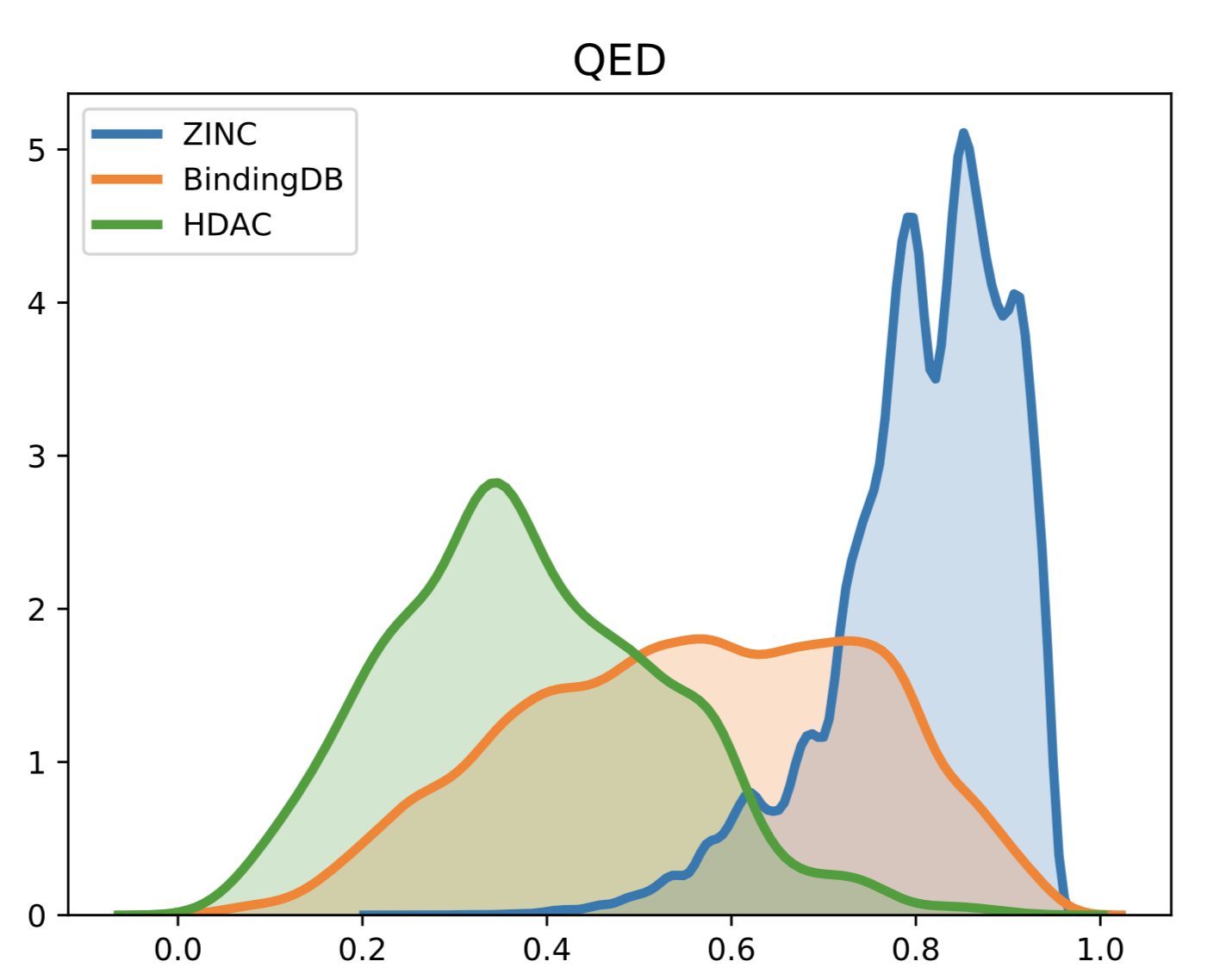}
    \caption{Distribution of QED for molecules in ZINC, BindingDB and for molecules that bind with HDAC1 (pIC50>6)}
    \label{fig: bindingdb_vae}
\end{figure}

As discussed in Section~\ref{sec:hdac}, we generated molecules that targets Human HDAC1. Figure~\ref{fig: bindingdb_vae} shows the distribution of QED for molecules that are present in ZINC, BindingDB and the subset of molecules from BindingDB that binds to HDAC1. We can see that there are very few molecules with high QED that binds to HDAC1.
Table \ref{tab:hdac1} shows that CogMol-generated molecules comprise a larger proportion of molecules satisfying high pIC50 and QED criteria, implying CogMol can discover novel and optimal molecules even in a low-data regime.  Docking analysis of the generated molecules and comparison with training molecules are shown in Table~\ref{tab:hdac1_docking}, showing the binding energy of the generated molecules are comparable with that of the training molecules.

\label{SI:hdca1}

\begin{table}[bt]
\centering
\caption{Number of designed molecules with desired attributes compared to BindingDB ligands for HDAC1.}
\scalebox{0.85}{
\begin{tabular}{c|cccc}
\toprule
{Dataset}   & {Total Molecules} & {QED\textgreater{}0.8} & {pIC50\textgreater{}6 and QED \textgreater 0.8} & {pIC50 \textgreater 7} and {QED \textgreater 0.8} \\
\midrule
Train Set & 2253 & 43 (1.9\%)  & 9 (0.39\%)  & 1(0.04\%) \\
Generated & 1388 & 188 (13.6\%) & 89 (6.42\%) & 32(2.3\%) \\
\bottomrule
\end{tabular}
}
\label{tab:hdac1}
\end{table}

\begin{table}[!ht]
\centering
\caption{HDAC1 docking results: average ($\pm$ standard deviation) binding free energy, minimum binding free energy, and fraction of generated molecules with binding free energy $<-6$ kcal/mol for molecules from the train set (with QED > 0.7) and generated molecules.}
\label{tab:hdac1_docking}
\begin{tabular}{r|
                S[table-number-alignment=center,separate-uncertainty,table-figures-uncertainty=1]
                S[table-number-alignment=center,table-format=1.1]
                S[table-number-alignment=center]}
\toprule
{Dataset} & {Mean (kcal/mol)} & {Min (kcal/mol)} & {Low Energy (\%)} \\
\midrule
{Train Set} & -6.9 \pm 0.6 & -8.2 & 92.5\\
{Generated} & -7.1 \pm 0.8 & -8.8 & 89.9\\
\bottomrule
\end{tabular}
\vspace{10pt}
\end{table}

\FloatBarrier



\end{document}

%% file: math_commands.tex
\usepackage{amsmath,amsfonts,bm}

\def\eqref#1{equation~\ref{#1}}

\def\1{\bm{1}}

\def\rva{{\mathbf{a}}}

\def\rvx{{\mathbf{x}}}

\def\rvz{{\mathbf{z}}}

\DeclareMathAlphabet{\mathsfit}{\encodingdefault}{\sfdefault}{m}{sl}
\SetMathAlphabet{\mathsfit}{bold}{\encodingdefault}{\sfdefault}{bx}{n}

\newcommand{\E}{\mathbb{E}}

\newcommand{\R}{\mathbb{R}}

\newcommand{\KL}{D_{\mathrm{KL}}}

\newcommand{\invivo}{\emph{in vivo} }
\newcommand{\invitro}{\emph{in vitro} }
\newcommand{\insilico}{\emph{in silico} }

%% file: tbl_tox_MT-DNN.tex
\begin{table}[ht!]
    \begin{center}
    \caption{
        Performance on toxicity prediction using MT-DNN for all 12 Tox21 tasks and ClinTox task (CT-TOX). The reported metrics are ROC AUC, accuracy, balanced accuracy, true negative rate, true positive rate, precision, recall, and the F1 score. Refer to Table~\ref{tbl:tox_RF} for a comparison with random forest, the MT-DNN achieves much better true positive, recall and F1 score with slight penalty on AUC.
    }%
    \label{tbl:tox_prediction} 
    \scalebox{0.85}{
        \begin{tabular}{r|
        S[table-format=1.2]
        S[table-format=1.2]
        S[table-format=1.2]
        S[table-format=1.2]
        S[table-format=1.2]
        S[table-format=1.2]
        S[table-format=1.2]
        S[table-format=1.2]
        }
            \toprule
            {Task}
            & {AUC}
            & {ACC}
            & {BAC}
            & {TN}
            & {TP}
            & {PR}
            & {RC}
            & {F1}
            \\
            \midrule
            NR-AR & 0.72 & 0.97 & 0.73 & 0.99 & 0.46 & 0.98 & 0.46 & 0.55  \\
            NR-Aromatase & 0.78 & 0.95 & 0.65 & 0.99 & 0.31 & 0.96 & 0.31 & 0.41  \\
            NR-PPAR-$\gamma$ & 0.75 & 0.97 & 0.61 & 0.99 & 0.23 & 0.96 & 0.23 & 0.31  \\
            SR-HSE & 0.75 & 0.94 & 0.61 & 0.99 & 0.23 & 0.94 & 0.23 & 0.32  \\
            NR-AR-LBD & 0.82 & 0.98 & 0.78 & 0.99 & 0.56 & 0.99 & 0.56 & 0.66  \\
            NR-ER & 0.69 & 0.88 & 0.64 & 0.96 & 0.32 & 0.89 & 0.32 & 0.40  \\
            SR-ARE & 0.78 & 0.86 & 0.67 & 0.95 & 0.39 & 0.88 & 0.39 & 0.47  \\
            SR-MMP & 0.87 & 0.89 & 0.75 & 0.96 & 0.54 & 0.93 & 0.54 & 0.62  \\
            NR-AhR & 0.85 & 0.91 & 0.73 & 0.96 & 0.50 & 0.93 & 0.50 & 0.55  \\
            NR-ER-LBD & 0.78 & 0.96 & 0.69 & 0.99 & 0.38 & 0.97 & 0.38 & 0.48  \\
            SR-ATAD5 & 0.75 & 0.96 & 0.61 & 0.99 & 0.22 & 0.97 & 0.22 & 0.32  \\
            SR-p53 & 0.79 & 0.94 & 0.64 & 0.99 & 0.29 & 0.96 & 0.29 & 0.40 \\
            CT-TOX & 0.79 & 0.92 & 0.61 & 0.97 & 0.25 & 0.84 & 0.25 & 0.31 \\[1pt]
            \hdashline
            \rule{0pt}{0.8\normalbaselineskip}  
            Average	& 0.78 & 0.93 & \textbf{0.67} & 0.98 & \textbf{0.36} & 0.94 & \textbf{0.36} & \textbf{0.45} \\
            \bottomrule
        \end{tabular}
    }
    \end{center}
\end{table}

%% file: tbl_tox_RF.tex
\begin{table}[htb]
    \begin{center}
    \caption{
        Performance on toxicity prediction for Random Forest on all 12 Tox21 tasks and ClinTox task (CT-TOX). The reported metrics are ROC AUC, accuracy, balanced accuracy, true negative rate, true positive rate, precision, recall, and the F1 score.
    }%
    \label{tbl:tox_RF} 
    \scalebox{0.85}{
        \begin{tabular}{r|
        S[table-format=1.2]
        S[table-format=1.2]
        S[table-format=1.2]
        S[table-format=1.2]
        S[table-format=1.2]
        S[table-format=1.2]
        S[table-format=1.2]
        S[table-format=1.2]
        }
            \toprule
            {Task}
            & {AUC}
            & {ACC}
            & {BAC}
            & {TN}
            & {TP}
            & {PR}
            & {RC}
            & {F1}
            \\
            \midrule 
            NR-AR & 0.78 & 0.98 & 0.73 & 0.99 & 0.46 & 0.99 & 0.46 & 0.60 \\
            NR-Aromatase & 0.81 & 0.96 & 0.60 & 0.99 & 0.21 & 0.98 & 0.21 & 0.32 \\
            NR-PPAR-gamma & 0.82 & 0.97 & 0.56 & 0.99 & 0.12 & 0.99 & 0.12 & 0.20 \\
            SR-HSE & 0.77 & 0.95 & 0.56 & 0.99 & 0.13 & 0.97 & 0.13 & 0.22 \\
            NR-AR-LBD & 0.85 & 0.98 & 0.77 & 0.99 & 0.54 & 0.99 & 0.54 & 0.66 \\
            NR-ER & 0.72 & 0.89 & 0.60 & 0.98 & 0.21 & 0.93 & 0.21 & 0.32 \\
            SR-ARE & 0.81 & 0.86 & 0.61 & 0.99 & 0.24 & 0.94 & 0.24 & 0.36 \\
            SR-MMP & 0.88 & 0.89 & 0.67 & 0.98 & 0.36 & 0.95 & 0.36 & 0.49 \\
            NR-AhR & 0.89 & 0.91 & 0.65 & 0.99 & 0.32 & 0.96 & 0.32 & 0.45 \\
            NR-ER-LBD & 0.80 & 0.96 & 0.65 & 0.99 & 0.31 & 0.99 & 0.31 & 0.45 \\
            SR-ATAD5 & 0.84 & 0.96 & 0.55 & 0.99 & 0.10 & 0.97 & 0.10 & 0.18 \\
            SR-p53 & 0.83 & 0.95 & 0.56 & 0.99 & 0.13 & 0.99 & 0.13 & 0.22 \\
            CT-TOX & 0.76 & 0.92 & 0.55 & 0.98 & 0.13 & 0.85 & 0.13 & 0.19 \\[1pt]
            \hdashline
            \rule{0pt}{0.8\normalbaselineskip}  
            Average & \textbf{0.81} & 0.94 & 0.62 & 0.99 & 0.25 & 0.96 & 0.25 & 0.36 \\
            \bottomrule
        \end{tabular}
    }
    \end{center}
\end{table}

%% file: tbl_tox_counts.tex
\begin{table}[bth]
    \begin{center}
    \caption{
        Proportion of molecules being predicted to have a number of toxic endpoints. There are 200k molecules for each of the targets NSP9, RBD, and Main Protease. While there are only 680 molecules in the FDA database. Note that there are negligible amount of molecules or metabolites having more than 10 toxic endpoints and thus they are omitted here.
    }%
    \label{tbl:tox_counts} 
    \scalebox{0.85}{
        \begin{tabular}{r|
        S[table-format=1.3]
        S[table-format=1.3]
        S[table-format=1.3]
        S[table-format=1.3]
        S[table-format=1.3]
        S[table-format=1.3]
        S[table-format=1.3]
        S[table-format=1.3]
        S[table-format=1.3]
        S[table-format=1.3]
        S[table-format=1.3]
        }
            \toprule
            {Targets}
            & {0}
            & {1}
            & {2}
            & {3}
            & {4}
            & {5}
            & {6}
            & {7}
            & {8}
            & {9}
            & {10}
            \\
            \midrule 
            \underline{Molecules} \\[0.5mm]
            NSP9 & 
            0.502 & 0.223 & 0.121 & 0.076 & 0.042 & 0.021 & 0.01 & 0.004 & 0.001 & 0.0 & 0.0 
            \\
            RBD & 
            0.505 & 0.223 & 0.121 & 0.075 & 0.041 & 0.021 & 0.009 & 0.004 & 0.001 & 0.0 & 0.0 
            \\
            Main Protease & 
            0.507 & 0.221 & 0.123 & 0.073 & 0.04 & 0.021 & 0.009 & 0.004 & 0.001 & 0.0 & 0.0 
            \\
            FDA & 
            0.569 & 0.157 & 0.1 & 0.056 & 0.038 & 0.034 & 0.022 & 0.013 & 0.006 & 0.001 & 0.001 
            \\[2mm]
            \underline{Metabolites} 
            \\[0.5mm]
            NSP9 & 
            0.642 & 0.191 & 0.075 & 0.038 & 0.023 & 0.014 & 0.009 & 0.005 & 0.002 & 0.0 & 0.0 
            \\
            RBD & 
            0.63 & 0.206 & 0.085 & 0.04 & 0.019 & 0.011 & 0.005 & 0.002 & 0.001 & 0.0 & 0.0 
            \\
            Main Protease & 
            0.652 & 0.192 & 0.071 & 0.038 & 0.021 & 0.011 & 0.008 & 0.004 & 0.002 & 0.0 & 0.0 
            \\
            FDA & 
            0.656 & 0.138 & 0.068 & 0.045 & 0.037 & 0.025 & 0.013 & 0.01 & 0.004 & 0.002 & 0.001 
            \\
            \bottomrule
        \end{tabular}
    }
    \end{center}
\end{table}

%% file: main.bbl
\begin{thebibliography}{10}

\bibitem{matthews2016omics}
H.~Matthews, J.~Hanison, and N.~Nirmalan, ````{O}mics''-informed drug and
  biomarker discovery: opportunities, challenges and future perspectives,''
  {\em Proteomes}, vol.~4, no.~3, p.~28, 2016.

\bibitem{polishchuk2013estimation}
P.~G. Polishchuk, T.~I. Madzhidov, and A.~Varnek, ``Estimation of the size of
  drug-like chemical space based on {GDB}-17 data,'' {\em Journal of
  Computer-Aided Molecular Design}, vol.~27, no.~8, pp.~675--679, 2013.

\bibitem{gomez2018automatic}
R.~G{\'o}mez-Bombarelli, J.~N. Wei, D.~Duvenaud, J.~M. Hern{\'a}ndez-Lobato,
  B.~S{\'a}nchez-Lengeling, D.~Sheberla, J.~Aguilera-Iparraguirre, T.~D.
  Hirzel, R.~P. Adams, and A.~Aspuru-Guzik, ``Automatic chemical design using a
  data-driven continuous representation of molecules,'' {\em ACS Central
  Science}, vol.~4, no.~2, pp.~268--276, 2018.

\bibitem{Zhavoronkov2019natbio}
A.~Zhavoronkov, Y.~A. Ivanenkov, A.~Aliper, M.~S. Veselov, V.~A. Aladinskiy,
  A.~V. Aladinskaya, V.~A. .~Terentiev, D.~A. Polykovskiy, M.~D. Kuznetsov,
  A.~Asadulaev, Y.~Volkov, A.~Zholus, R.~R. Shayakhmetov, A.~Zhebrak, L.~I.
  Minaeva, B.~A. Zagribelnyy, L.~H. Lee, R.~Soll, D.~Madge, L.~Xing, G.~Tao,
  and A.~Aspuru-Guzik, ``Deep learning enables rapid identification of potent
  {DDR1} kinase inhibitors,'' {\em Nature Biotechnology}, vol.~37,
  pp.~1038--–1040, 2019.

\bibitem{zhou2019optimization}
Z.~Zhou, S.~Kearnes, L.~Li, R.~N. Zare, and P.~Riley, ``Optimization of
  molecules via deep reinforcement learning,'' {\em Scientific Reports},
  vol.~9, no.~1, p.~10752, 2019.

\bibitem{peon2017predicting}
A.~Pe{\'o}n, S.~Naulaerts, and P.~J. Ballester, ``Predicting the reliability of
  drug-target interaction predictions with maximum coverage of target space,''
  {\em Scientific reports}, vol.~7, no.~1, pp.~1--11, 2017.

\bibitem{cheng2019network}
F.~Cheng, I.~A. Kov{\'a}cs, and A.-L. Barab{\'a}si, ``Network-based prediction
  of drug combinations,'' {\em Nature communications}, vol.~10, no.~1,
  pp.~1--11, 2019.

\bibitem{miljkovic2018data}
F.~Miljkovi{\'c} and J.~Bajorath, ``Data-driven exploration of selectivity and
  off-target activities of designated chemical probes,'' {\em Molecules},
  vol.~23, no.~10, p.~2434, 2018.

\bibitem{Alley_2019}
E.~C. Alley, G.~Khimulya, S.~Biswas, M.~AlQuraishi, and G.~M. Church, ``Unified
  rational protein engineering with sequence-based deep representation
  learning,'' {\em Nature Methods}, vol.~16, p.~1315–1322, Oct 2019.

\bibitem{rao2019evaluating}
R.~Rao, N.~Bhattacharya, N.~Thomas, Y.~Duan, P.~Chen, J.~Canny, P.~Abbeel, and
  Y.~Song, ``Evaluating protein transfer learning with {TAPE},'' in {\em
  Advances in Neural Information Processing Systems}, pp.~9686--9698, 2019.

\bibitem{segler2017generating}
M.~H. Segler, T.~Kogej, C.~Tyrchan, and M.~P. Waller, ``Generating focused
  molecule libraries for drug discovery with recurrent neural networks,'' {\em
  ACS Central Science}, vol.~4, no.~1, pp.~120--131, 2017.

\bibitem{gupta2018generative}
A.~Gupta, A.~T. M{\"u}ller, B.~J. Huisman, J.~A. Fuchs, P.~Schneider, and
  G.~Schneider, ``Generative recurrent networks for de novo drug design,'' {\em
  Molecular Informatics}, vol.~37, no.~1-2, p.~1700111, 2018.

\bibitem{blaschke2018application}
T.~Blaschke, M.~Olivecrona, O.~Engkvist, J.~Bajorath, and H.~Chen,
  ``Application of generative autoencoder in de novo molecular design,'' {\em
  Molecular Informatics}, vol.~37, no.~1-2, p.~1700123, 2018.

\bibitem{kang2018conditional}
S.~Kang and K.~Cho, ``Conditional molecular design with deep generative
  models,'' {\em Journal of Chemical Information and Modeling}, vol.~59, no.~1,
  pp.~43--52, 2018.

\bibitem{lim2018molecular}
J.~Lim, S.~Ryu, J.~W. Kim, and W.~Y. Kim, ``Molecular generative model based on
  conditional variational autoencoder for de novo molecular design,'' {\em
  Journal of Cheminformatics}, vol.~10, no.~1, p.~31, 2018.

\bibitem{guimaraes2017objective}
G.~L. Guimaraes, B.~Sanchez-Lengeling, C.~Outeiral, P.~L.~C. Farias, and
  A.~Aspuru-Guzik, ``Objective-reinforced generative adversarial networks
  ({ORGAN}) for sequence generation models,'' {\em arXiv preprint
  arXiv:1705.10843}, 2017.

\bibitem{de2018molgan}
N.~De~Cao and T.~Kipf, ``{MolGAN}: An implicit generative model for small
  molecular graphs,'' {\em arXiv preprint arXiv:1805.11973}, 2018.

\bibitem{kusner2017grammar}
M.~J. Kusner, B.~Paige, and J.~M. Hern{\'a}ndez-Lobato, ``Grammar variational
  autoencoder,'' in {\em Proceedings of the 34th International Conference on
  Machine Learning}, pp.~1945--1954, JMLR.org, 2017.

\bibitem{dai2018syntax}
H.~Dai, Y.~Tian, B.~Dai, S.~Skiena, and L.~Song, ``Syntax-directed variational
  autoencoder for structured data,'' {\em arXiv preprint arXiv:1802.08786},
  2018.

\bibitem{li2018learning}
Y.~Li, O.~Vinyals, C.~Dyer, R.~Pascanu, and P.~Battaglia, ``Learning deep
  generative models of graphs,'' {\em arXiv preprint arXiv:1803.03324}, 2018.

\bibitem{li2018multi}
Y.~Li, L.~Zhang, and Z.~Liu, ``Multi-objective de novo drug design with
  conditional graph generative model,'' {\em Journal of Cheminformatics},
  vol.~10, no.~1, p.~33, 2018.

\bibitem{simonovsky2018graphvae}
M.~Simonovsky and N.~Komodakis, ``{GraphVAE}: Towards generation of small
  graphs using variational autoencoders,'' in {\em International Conference on
  Artificial Neural Networks}, pp.~412--422, Springer, 2018.

\bibitem{samanta2019nevae}
B.~Samanta, D.~Abir, G.~Jana, P.~K. Chattaraj, N.~Ganguly, and M.~G. Rodriguez,
  ``{NeVAE}: a deep generative model for molecular graphs,'' in {\em
  Proceedings of the AAAI Conference on Artificial Intelligence}, vol.~33,
  pp.~1110--1117, 2019.

\bibitem{ma2018constrained}
T.~Ma, J.~Chen, and C.~Xiao, ``Constrained generation of semantically valid
  graphs via regularizing variational autoencoders,'' in {\em Advances in
  Neural Information Processing Systems}, pp.~7113--7124, 2018.

\bibitem{kajino2018molecular}
H.~Kajino, ``Molecular hypergraph grammar with its application to molecular
  optimization,'' {\em arXiv preprint arXiv:1809.02745}, 2018.

\bibitem{jin2018junction}
W.~Jin, R.~Barzilay, and T.~Jaakkola, ``Junction tree variational autoencoder
  for molecular graph generation,'' {\em arXiv preprint arXiv:1802.04364},
  2018.

\bibitem{popova2018deep}
M.~Popova, O.~Isayev, and A.~Tropsha, ``Deep reinforcement learning for de novo
  drug design,'' {\em Science Advances}, vol.~4, no.~7, p.~eaap7885, 2018.

\bibitem{olivecrona2017molecular}
M.~Olivecrona, T.~Blaschke, O.~Engkvist, and H.~Chen, ``Molecular de-novo
  design through deep reinforcement learning,'' {\em Journal of
  Cheminformatics}, vol.~9, no.~1, p.~48, 2017.

\bibitem{jaques2017sequence}
N.~Jaques, S.~Gu, D.~Bahdanau, J.~M. Hern{\'a}ndez-Lobato, R.~E. Turner, and
  D.~Eck, ``Sequence tutor: Conservative fine-tuning of sequence generation
  models with {KL}-control,'' in {\em Proceedings of the 34th International
  Conference on Machine Learning}, pp.~1645--1654, JMLR.org, 2017.

\bibitem{putin2018reinforced}
E.~Putin, A.~Asadulaev, Y.~Ivanenkov, V.~Aladinskiy, B.~Sanchez-Lengeling,
  A.~Aspuru-Guzik, and A.~Zhavoronkov, ``Reinforced adversarial neural computer
  for de novo molecular design,'' {\em Journal of Chemical Information and
  Modeling}, vol.~58, no.~6, pp.~1194--1204, 2018.

\bibitem{born2020paccmann}
J.~Born, M.~Manica, A.~Oskooei, J.~Cadow, and M.~R. Mart{\'\i}nez,
  ``{PaccMann\textsuperscript{RL}}: Designing anticancer drugs from
  transcriptomic data via reinforcement learning,'' in {\em International
  Conference on Research in Computational Molecular Biology}, pp.~231--233,
  Springer, 2020.

\bibitem{das2020science}
P.~Das, T.~Sercu, K.~Wadhawan, I.~Padhi, S.~Gehrmann, F.~Cipcigan,
  V.~Chenthamarakshan, H.~Strobelt, C.~d. Santos, P.-Y. Chen, {\em et~al.},
  ``Accelerating antimicrobial discovery with controllable deep generative
  models and molecular dynamics,'' {\em arXiv preprint arXiv:2005.11248}, 2020.

\bibitem{skalic2019target}
M.~Skalic, D.~Sabbadin, B.~Sattarov, S.~Sciabola, and G.~De~Fabritiis, ``From
  target to drug: Generative modeling for the multimodal structure-based ligand
  design,'' {\em Molecular pharmaceutics}, vol.~16, no.~10, pp.~4282--4291,
  2019.

\bibitem{grechishnikova2019transformer}
D.~A. Grechishnikova, ``Transformer neural network for protein specific de novo
  drug generation as machine translation problem,'' {\em BioRxiv}, p.~863415,
  2019.

\bibitem{polykovskiy2018molecular}
D.~Polykovskiy, A.~Zhebrak, B.~Sanchez-Lengeling, S.~Golovanov, O.~Tatanov,
  S.~Belyaev, R.~Kurbanov, A.~Artamonov, V.~Aladinskiy, M.~Veselov, A.~Kadurin,
  S.~Nikolenko, A.~Aspuru-Guzik, and A.~Zhavoronkov, ``Molecular sets
  ({MOSES}): A benchmarking platform for molecular generation models,'' {\em
  arXiv preprint arXiv:1811.12823}, 2018.

\bibitem{irwin2005zinc}
J.~J. Irwin and B.~K. Shoichet, ``{ZINC}--a free database of commercially
  available compounds for virtual screening,'' {\em Journal of Chemical
  Information and Modeling}, vol.~45, no.~1, pp.~177--182, 2005.

\bibitem{gilson2015bindingdb}
M.~K. Gilson, T.~Liu, M.~Baitaluk, G.~Nicola, L.~Hwang, and J.~Chong,
  ``{BindingDB} in 2015: a public database for medicinal chemistry,
  computational chemistry and systems pharmacology,'' {\em Nucleic Acids
  Research}, vol.~44, no.~D1, pp.~D1045--D1053, 2015.

\bibitem{karimi2018deepaffinity}
M.~Karimi, D.~Wu, Z.~Wang, and Y.~Shen, ``{DeepAffinity}: interpretable deep
  learning of compound-protein affinity through unified recurrent and
  convolutional neural networks,'' {\em arXiv preprint arXiv:1806.07537}, 2018.

\bibitem{kingma2013auto}
D.~P. Kingma and M.~Welling, ``Auto-encoding variational {B}ayes,'' {\em arXiv
  preprint arXiv:1312.6114}, 2013.

\bibitem{bosc2017use}
N.~Bosc, C.~Meyer, and P.~Bonnet, ``The use of novel selectivity metrics in
  kinase research,'' {\em BMC Bioinformatics}, vol.~18, no.~1, p.~17, 2017.

\bibitem{Tox21}
R.~Huang, M.~Xia, D.-T. Nguyen, T.~Zhao, S.~Sakamuru, J.~Zhao, S.~A. Shahane,
  A.~Rossoshek, and A.~Simeonov, ``{Tox21} challenge to build predictive models
  of nuclear receptor and stress response pathways as mediated by exposure to
  environmental chemicals and drugs,'' {\em Frontiers in Environmental
  Science}, vol.~3, p.~85, 2016.

\bibitem{Wu2018}
Z.~Wu, B.~Ramsundar, E.~N. Feinberg, J.~Gomes, C.~Geniesse, A.~S. Pappu,
  K.~Leswing, and V.~Pande, ``Molecule{N}et: a benchmark for molecular machine
  learning,'' {\em Chemical Science}, vol.~9, no.~2, pp.~513--530, 2018.

\bibitem{trott2010autodock}
O.~Trott and A.~J. Olson, ``Auto{D}ock {V}ina: improving the speed and accuracy
  of docking with a new scoring function, efficient optimization, and
  multithreading,'' {\em Journal of Computational Chemistry}, vol.~31, no.~2,
  pp.~455--461, 2010.

\bibitem{schwaller2020predicting}
P.~Schwaller, R.~Petraglia, V.~Zullo, V.~H. Nair, R.~A. Haeuselmann, R.~Pisoni,
  C.~Bekas, A.~Iuliano, and T.~Laino, ``Predicting retrosynthetic pathways
  using transformer-based models and a hyper-graph exploration strategy,'' {\em
  Chemical Science}, vol.~11, no.~12, pp.~3316--3325, 2020.

\bibitem{SchwallerFWD}
P.~Schwaller, T.~Laino, T.~Gaudin, P.~Bolgar, C.~A. Hunter, C.~Bekas, and A.~A.
  Lee, ``Molecular transformer: A model for uncertainty-calibrated chemical
  reaction prediction,'' {\em ACS Central Science}, vol.~5, no.~9,
  pp.~1572--1583, 2019.

\bibitem{preuer2018frechet}
K.~Preuer, P.~Renz, T.~Unterthiner, S.~Hochreiter, and G.~Klambauer,
  ``{Fr{\'e}chet ChemNet} distance: a metric for generative models for
  molecules in drug discovery,'' {\em Journal of Chemical Information and
  Modeling}, vol.~58, no.~9, pp.~1736--1741, 2018.

\bibitem{tanimoto_willett}
P.~Willett, J.~M. Barnard, and G.~M. Downs, ``Chemical similarity searching,''
  {\em Journal of Chemical Information and Computer Sciences}, vol.~38, no.~6,
  pp.~983--996, 1998.

\bibitem{pubchem_kim}
S.~Kim, J.~Chen, T.~Cheng, A.~Gindulyte, J.~He, S.~He, Q.~Li, B.~A. Shoemaker,
  P.~A. Thiessen, B.~Yu, L.~Zaslavsky, J.~Zhang, and E.~E. Bolton, ``{{PubChem}
  2019 update: improved access to chemical data},'' {\em Nucleic Acids
  Research}, vol.~47, pp.~D1102--D1109, 10 2018.

\bibitem{hoffmann2020sars}
M.~Hoffmann, H.~Kleine-Weber, S.~Schroeder, N.~Kr{\"u}ger, T.~Herrler,
  S.~Erichsen, T.~S. Schiergens, G.~Herrler, N.-H. Wu, A.~Nitsche, {\em
  et~al.}, ``{SARS-CoV-2} cell entry depends on {ACE2} and {TMPRSS2} and is
  blocked by a clinically proven protease inhibitor,'' {\em Cell}, 2020.

\bibitem{liu2020hydroxychloroquine}
J.~Liu, R.~Cao, M.~Xu, X.~Wang, H.~Zhang, H.~Hu, Y.~Li, Z.~Hu, W.~Zhong, and
  M.~Wang, ``Hydroxychloroquine, a less toxic derivative of chloroquine, is
  effective in inhibiting {SARS-CoV-2} infection in vitro,'' {\em Cell
  Discovery}, vol.~6, no.~1, pp.~1--4, 2020.

\bibitem{cao2008maximum}
Y.~Cao, T.~Jiang, and T.~Girke, ``A maximum common substructure-based algorithm
  for searching and predicting drug-like compounds,'' {\em Bioinformatics},
  vol.~24, no.~13, pp.~i366--i374, 2008.

\bibitem{FDAenamine}
EnamineStore, {\em FDA approved Drugs}, 2020 (accessed May, 2020).

\bibitem{litsa_das_kavraki_2020}
E.~Litsa, P.~Das, and L.~Kavraki, ``Prediction of drug metabolites using neural
  machine translation,'' May 2020.

\bibitem{fernandes2020economic}
N.~Fernandes, ``Economic effects of coronavirus outbreak {(COVID-19)} on the
  world economy,'' {\em Available at SSRN 3557504}, 2020.

\bibitem{agostini2018coronavirus}
M.~L. Agostini, E.~L. Andres, A.~C. Sims, R.~L. Graham, T.~P. Sheahan, X.~Lu,
  E.~C. Smith, J.~B. Case, J.~Y. Feng, R.~Jordan, {\em et~al.}, ``Coronavirus
  susceptibility to the antiviral remdesivir {(GS-5734)} is mediated by the
  viral polymerase and the proofreading exoribonuclease,'' {\em MBio}, vol.~9,
  no.~2, pp.~e00221--18, 2018.

\bibitem{le2020covid}
T.~T. Le, Z.~Andreadakis, A.~Kumar, R.~G. Roman, S.~Tollefsen, M.~Saville, and
  S.~Mayhew, ``The {COVID-19} vaccine development landscape,'' {\em Nat Rev
  Drug Discov}, vol.~19, no.~5, pp.~305--6, 2020.

\bibitem{wang2020remdesivir}
Y.~Wang, D.~Zhang, G.~Du, R.~Du, J.~Zhao, Y.~Jin, S.~Fu, L.~Gao, Z.~Cheng,
  Q.~Lu, {\em et~al.}, ``Remdesivir in adults with severe {COVID-19}: a
  randomised, double-blind, placebo-controlled, multicentre trial,'' {\em The
  Lancet}, 2020.

\bibitem{maccs_durant}
J.~L. Durant, B.~A. Leland, D.~R. Henry, and J.~G. Nourse, ``Reoptimization of
  {MDL} keys for use in drug discovery,'' {\em Journal of Chemical Information
  and Computer Sciences}, vol.~42, no.~6, pp.~1273--1280, 2002.

\bibitem{Srivastava2014}
N.~Srivastava, G.~Hinton, A.~Krizhevsky, I.~Sutskever, and R.~Salakhutdinov,
  ``Dropout: a simple way to prevent neural networks from overfitting,'' {\em
  The Journal of Machine Learning Research}, vol.~15, no.~1, pp.~1929--1958,
  2014.

\bibitem{Rogers2010}
D.~Rogers and M.~Hahn, ``Extended-connectivity fingerprints,'' {\em Journal of
  Chemical Information and Modeling}, vol.~50, no.~5, pp.~742--754, 2010.

\bibitem{Liu2019}
S.~Liu, M.~F. Demirel, and Y.~Liang, ``N-gram graph: Simple unsupervised
  representation for graphs, with applications to molecules,'' in {\em Advances
  in Neural Information Processing Systems}, pp.~8464--8476, 2019.

\bibitem{zhang2020crystal}
L.~Zhang, D.~Lin, X.~Sun, U.~Curth, C.~Drosten, L.~Sauerhering, S.~Becker,
  K.~Rox, and R.~Hilgenfeld, ``Crystal structure of {SARS-CoV-2} main protease
  provides a basis for design of improved $\alpha$-ketoamide inhibitors,'' {\em
  Science}, vol.~368, no.~6489, pp.~409--412, 2020.

\bibitem{krivak2018p2rank}
R.~Kriv{\'a}k and D.~Hoksza, ``{P2Rank}: machine learning based tool for rapid
  and accurate prediction of ligand binding sites from protein structure,''
  {\em Journal of Cheminformatics}, vol.~10, no.~1, p.~39, 2018.

\bibitem{jendele2019prankweb}
L.~Jendele, R.~Krivak, P.~Skoda, M.~Novotny, and D.~Hoksza, ``Prank{W}eb: a web
  server for ligand binding site prediction and visualization,'' {\em Nucleic
  Acids Research}, vol.~47, no.~W1, pp.~W345--W349, 2019.

\bibitem{eMolecules}
eMolecules, {\em {eMolecules} Plus Database Download}, 2020 (accessed May,
  2020).

\end{thebibliography}
